\documentclass[12pt]{article}

\usepackage{newtxtext,newtxmath}

\usepackage{graphicx}

\usepackage[letterpaper,margin=1in]{geometry}

\renewenvironment{abstract}
	{\quotation}
	{\endquotation}

\date{}

\makeatletter
\renewcommand{\fnum@figure}{\textbf{Figure \thefigure}}
\renewcommand{\fnum@table}{\textbf{Table \thetable}}
\makeatother

\usepackage{scicite}

\usepackage{url}
\usepackage{algorithm}
\usepackage{algpseudocode}

\def\scititle{
	High-speed control and navigation for quadrupedal robots on complex and discrete terrain
}

\title{\bfseries \boldmath \scititle}

\author{
	Hyeongjun Kim, Hyunsik Oh, Jeongsoo Park,\\
    Yunho Kim, Donghoon Youm, Moonkyu Jung, \\
    Minho Lee, Jemin Hwangbo$^{\ast}$\and
	\small Robotics \& Artificial Intelligence Lab, KAIST, Daejeon, Korea \and
	\small$^\ast$Corresponding author. Email: jhwangbo@kaist.ac.kr\and
}

\begin{document} 
\maketitle
\begin{abstract} \bfseries \boldmath
High-speed legged navigation in discrete and geometrically complex environments is a challenging task because of the high–degree-of-freedom dynamics and long-horizon, nonconvex nature of the optimization problem. In this work, we propose a hierarchical navigation pipeline for legged robots that can traverse such environments at high speed. The proposed pipeline consists of a planner and tracker module. The planner module finds physically feasible foothold plans by sampling-based optimization with fast sequential filtering using heuristics and a neural network. Subsequently, rollouts are performed in a physics simulation to identify the best foothold plan regarding the engineered cost function and to confirm its physical consistency. This hierarchical planning module is computationally efficient and physically accurate at the same time. The tracker aims to accurately step on the target footholds from the planning module. During the training stage, the foothold target distribution is given by a generative model that is trained competitively with the tracker. This process ensures that the tracker is trained in an environment with the desired difficulty. The resulting tracker can overcome terrains that are more difficult than what the previous methods could manage. We demonstrated our approach using Raibo, our in-house dynamic quadruped robot. The results were dynamic and agile motions: Raibo is capable of running on vertical walls, jumping a 1.3-meter gap, running over stepping stones at 4 meters per second, and autonomously navigating on terrains full of 30° ramps, stairs, and boxes of various sizes.
\end{abstract}

\paragraph{One-Sentence Summary:}
A framework for legged navigation, which enables high-speed running on complex and discrete terrain.

\noindent
\section*{Introduction}
With recent advances in robotic technology, there has been an increase in efforts to replace humans with robots in certain workplaces. In particular, legged robots are promising candidates to replace humans in search and rescue missions at disaster sites and construction areas because of their ability to efficiently traverse various challenging terrains. Rapid exploration and extensive area coverage are paramount for these missions. However, rapid locomotion in such environments, which consist of discontinuous terrains such as stairs, steps, large debris, and gaps, is still challenging for legged robots. These tasks require not only a controller capable of generating highly dynamic motions but also a fast and dynamically consistent navigation algorithm, both of which are still challenging to develop. To swiftly navigate discontinuous terrains, finding a feasible foothold plan for the given environment and coordinating all joint actuators to track the foothold plan precisely are crucial. However, this task presents several challenges. First, finding a feasible foothold plan on discontinuous terrains is difficult because of the nonconvexity of the optimization problem. Most optimization algorithms exploit the gradient to narrow the search space, assuming the convexity of the problem. Without such a structure of the problem, multiple initializations or random sampling must be performed. However, these approaches are prone to the curse of dimensionality and often become computationally intractable. Second, given frequent acceleration and deceleration during agile movements, a longer time horizon has to be considered to find a dynamically feasible path. Given the exponentially increasing number of possible paths in time, the necessary computation might be unmanageably high. In addition, because the robot cannot change directions so quickly at high speed, the full dynamics must be considered to evaluate whether a path is achievable. This further increases the computational burden, potentially exceeding the capabilities of on-board processing. Last, accurate stepping on the target foothold during high-speed movements poses a substantial control challenge. A slight delay in actuation can lead to completely different footholds given the rapid leg motion. In addition, the underactuated degrees of freedom of the base make closed-loop control less effective. Therefore, accurate modeling, a high-frequency control loop, and minimal control delay are desirable. Reinforcement learning (RL) is known to exhibit some of these characteristics. However, it also fails without a proper learning curriculum if the target environment distribution is highly challenging. Defining an effective curriculum for such arbitrary environments is another conundrum.
\subsubsection*{Related work}
Legged locomotion has remained a prominent research topic over the past decades. Advancements in hardware \cite{katz2019mini,hutter2016anymal,shin2022design,semini2011design} and control methodologies have enabled applications not only on flat terrains but also in various challenging environments \cite{miki2022learning,shi2024rethinking,miki2024learning,jenelten2022tamols,abdalla2023efficient}. In terms of control, two recent approaches stand out: optimization-based control (OC) and reinforcement learning (RL).

OC computes optimal control actions on the basis of dynamic models and has demonstrated strong performance in various gaits \cite{di2018dynamic}, high-speed locomotion \cite{kim2019highly}, push recovery \cite{chen2023quadruped}, and robust control under unknown payloads \cite{minniti2021adaptive}, as well as in other recent works \cite{li2024cafe,garcia2021time,hong2020real}. Its ability to incorporate constraints allows precise control of position and velocity, enabling accurate foot placements on discrete terrains such as gaps and stepping stones \cite{jenelten2022tamols,nguyen2022continuous,grandia2023perceptive,agrawal2022vision}.
However, optimizing a dynamically consistent plan in real time for high-degree-of-freedom robots is computationally intensive. Simplifications, such as fixed contact timing or approximated dynamics are often required but may degrade performance or lead to failure, especially during high-speed motions.

In contrast, RL performs optimization during the training phase in simulation and directly executes the learned policy at deployment. It adapts to a wide range of environmental and robot variations, such as compliance \cite{choi2023learning}, friction \cite{ji2022concurrent}, base mass \cite{kumar2021rma}, density \cite{feng2023genloco}, joint damping \cite{peng2018sim}, and motor friction \cite{kim2024constraint}. Recent works \cite{jenelten2024dtc,hoeller2024anymal,cheng2023extreme,duan2022learning} have shown that RL can achieve precise foot placements and natural motions even on challenging discrete terrains such as stairs, gaps, and stepping stones.
RL is a process where an agent improves its policy through interactions with the environment and feedback from rewards. However, if the environment is too easy or difficult relative to the agent's current performance, it becomes hard to observe meaningful reward differences, leading to stagnation in learning.
To address this issue, various curriculum-learning methods have been proposed. The simplest approach is to linearly increase the difficulty of the environment, such as the size of obstacles \cite{rudin2022advanced,jeon2023learning}.
However, these methods do not consider the agent’s performance or the robot’s physical limits, often leading to overly difficult or trivial environments that fail to provide meaningful reward signals.
To address this issue, two main approaches have been proposed: one stores feasible environment parameters as particles and evolves them over time \cite{wang2019,lee2020}, while the other discretizes the environment parameter space into grids and adjusts the sampling ratio within each grid \cite{xie2020allsteps}. Both methods enable adaptive environment generation, but they face limitations due to the curse of dimensionality when the number of environment parameters increases.

Conventional navigation systems typically adopt a hierarchical structure composed of a high-level module that generates suitable commands and a low-level module that controls the robot to follow these commands \cite{lee2024wheelnavi, zhang2024resilient, kim2022learning, chestnutt2007navigation, wellhausen2021rough}. The high-level module is generally implemented using either sampling-based or learning-based approaches.

Sampling-based methods randomly explore the search space to find feasible paths and are robust to sparse gradient issues. However, in high-degree-of-freedom legged robots, the computational cost becomes substantial. To mitigate this, methods have been proposed to approximate the robot’s morphology using reachability volumes \cite{wellhausen2021rough} or to balance sampling quality and computation time \cite{xu2021contact}. Nonetheless, these approaches often neglect dynamics, resulting in a lack of dynamic movement.
Learning-based high-level modules \cite{tsounis2020deepgait,heess2017emergence} aim to select appropriate locomotion skills, such as walking or jumping \cite{hoeller2024anymal,caluwaerts2023barkour}, or to generate foothold plans and base motion trajectories \cite{peng2017deeploco,brakel2022learning}. For instance, in \cite{brakel2022learning}, trajectory optimization was first performed on pre-defined maps to obtain feasible footholds, which are then used to train a network via imitation learning. However, such learning-based approaches tend to show limited generalization to unseen environments and struggle with long-horizon planning, limiting their performance in geometrically complex terrains.

\subsubsection*{Contributions}
The contributions of our work are as follows. We present a generative model, referred to as the map generator, which evolves competitively with the controller to enhance the robot’s athletic capabilities notably. In conjunction with this, we propose a controller capable of executing extreme parkour maneuvers, including jumping a 1.3-m gap and running on vertical walls. In addition, we introduce a foothold planning module that efficiently generates dynamically consistent foothold plans. We validate our approach on the Raibo robot, which demonstrates the ability to run on walls, achieve speeds of up to 4 m/s on stepping stones, perform consecutive jumps on tilted pads, and plan efficient routes across discrete and randomly generated terrains in real-time (Fig.~\ref{fig:scenarios} and Movie 1).

\section*{Results}
We developed a tracker module that could perform agile parkour motions and a sampling-based planner that plans efficient routes through geometrically complex environments in real time (Fig.~\ref{fig:pipeline}). By integrating these two modules, we developed a system that was capable of navigating challenging, discrete terrains at high speeds. The performance of the proposed system was validated through both simulations and real-world experiments.

\subsubsection*{Experimental setup}
Raibo \cite{choi2023learning}, our in-house dynamic quadrupedal robot, was used in all hardware experiments in this work. It weighs around 27.4 kg and has dimensions of 72 cm by 38.5 cm by 49 cm. Each joint actuator can produce a torque of up to 60 Nm and a rotation speed of up to 37.4 rad/s. The lengths of the thigh and shank are 24 cm each, making the total leg length 48 cm. Raibo’s onboard computer is a single 13th-generation Next Unit of Computing (NUC) computer with 12 cores and 16 threads. During real-world experiments, both the planner and tracker modules ran on the onboard computer.

In this study, two technical limitations precluded the use of a perception system. First, the high-speed movements, reaching up to 4 m/s, resulted in high accelerations. The positional drift in our state estimation algorithm made controlling the robot nearly impossible. Second, many desired footholds were either occluded or out of sight from our camera setup. In this case, a well-trained agent tends to make safe and slow motions when navigating in unknown terrains.

We believed these limitations could be addressed by improved state estimation and mapping algorithms and the use of multiple wide-angle cameras. However, this is beyond the scope of this work. Therefore, instead of using a perception system, we pre-acquired the height map information and obtained the robot's pose from a motion capture system using Vicon's Vero device. The robot’s pose was retrieved only at the moment when the planner module was activated.

\begin{figure}[p]
\includegraphics[width=\linewidth]{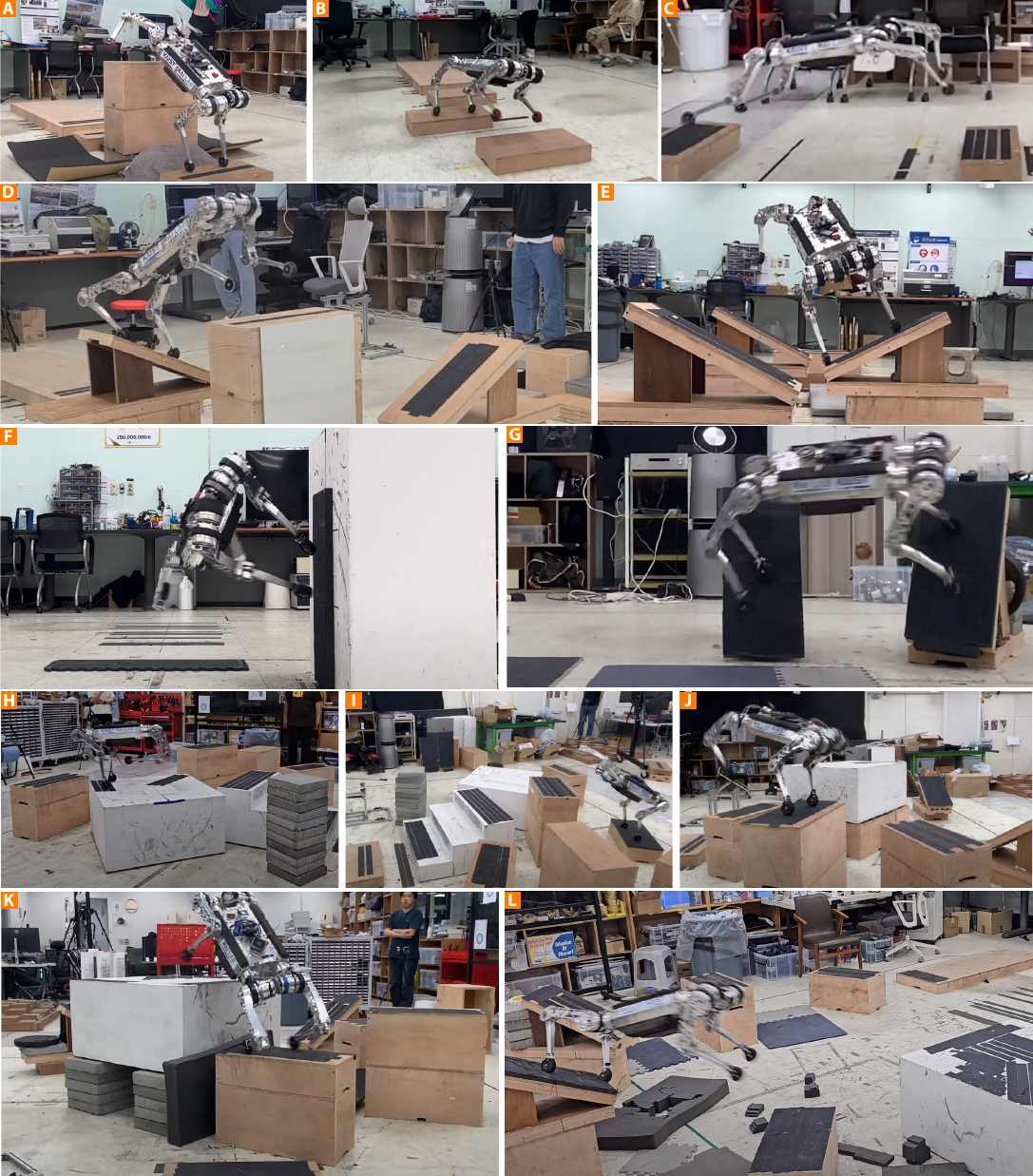}
\caption{\textbf{Raibo on various experiment scenarios.} Raibo performs extreme parkour maneuvers and high-speed navigation in various scenarios: (\textbf{A}) Ascent descent, (\textbf{B}) Side flutter, (\textbf{C}) 1.3 meter gap, (\textbf{D}) Double ramp, (\textbf{E}) Triple ramp, (\textbf{F}) Mono wall, (\textbf{G}) Double wall, (\textbf{H}) Stepscape\_left, (\textbf{I}) Stepscape\_right, (\textbf{J}) Ringnest\_orbit, (\textbf{K}) Ringnest\_corecut, (\textbf{L}) Flipturn. }
\label{fig:scenarios}
\end{figure}

Fig.~\ref{fig:scenarios} shows photos of some of the real experiments conducted within this area.
Each photo represents a scenario with a different terrain or experimental setting:
Fig.~\ref{fig:scenarios}(A–G) displays scenarios designed to demonstrate the tracker's performance.
Fig.~\ref{fig:scenarios}(H–L) displays scenarios designed to demonstrate the combined performance of the tracker and planner. As shown in the simulation in Movie 1, the areas of the ground other than the stepping stones were notably sunken, making it impossible for the robot to step.

Scenario names are indicated in the legend of Fig.~\ref{fig:scenarios}, with supplementary descriptions provided below for those requiring further clarification.
In the following text, each scenario in Fig.~\ref{fig:scenarios} will be referred to by its name in the figure legend.
“Ascent descent” is a scenario requiring ascent and descent of a 0.6-m obstacle. “1.3-m gap” is a scenario requiring a jump across a 1.3-m-wide gap, showcasing the robot’s ability to leap across a distance approximately 2.7 times its leg length. This surpasses the 1-m (1.4 times the leg length)gap crossed by Anymal D (27) and the 0.8-m (2 times the leg length)gap crossed by Unitree-A1 (28). “Double ramp” and “triple ramp” overcame consecutive ramps with 25°. “Mono wall” and “double wall” scenarios include terrain features that pose notable challenges for existing control methods. In particular, double wall required the robot to cross a 2.5-m gap, which was nearly impassable without using the walls as intermediate footholds. “Stepscape\_left” and “stepscape\_right” are experiments conducted on a terrain that included stairs, a pillar with a 0.3-m width and length, 0.3 m–by–0.6 m pillars of various heights, and a 0.9 m–by–0.9 m–by–0.6 m box, with different goal positions (left and right) used in each case. “Ringnest\_orbit” and “ringnest\_corecut” demonstrated the emergence of diverse paths, such as detour and direct routes, under different cost functions, which served as evaluation criteria for selecting the optimal foothold plan when the same map and goal location were given. The map included pillars of varying heights and sloped surfaces. “Flipturn” demonstrated the pipeline’s ability to generate optimal foothold plans, even in environments that required rapid turning to reach the goal position. This map featured a sparse distribution of 0.3 m–by–0.6 m stepping stones, spaced 1.0 to 1.3 m apart, and 30° ramps positioned at turning points to induce centripetal forces.

\subsubsection*{Evaluation of tracker}

The performance of the tracker in simulation is presented in Fig.~\ref{fig:tracker}A. The hardware limits of the Raibo robot used in the real experiments, such as max torque and joint speed limits, were implemented in the simulation as well. Corresponding video clips can be found in movie S1.
The capability to overcome gaps or high obstacles that were substantially larger than the robot's dimensions and run on multiple walls demonstrated a level of control performance that was not commonly observed in previous studies.

\begin{figure}[p]
\includegraphics[width=\linewidth]{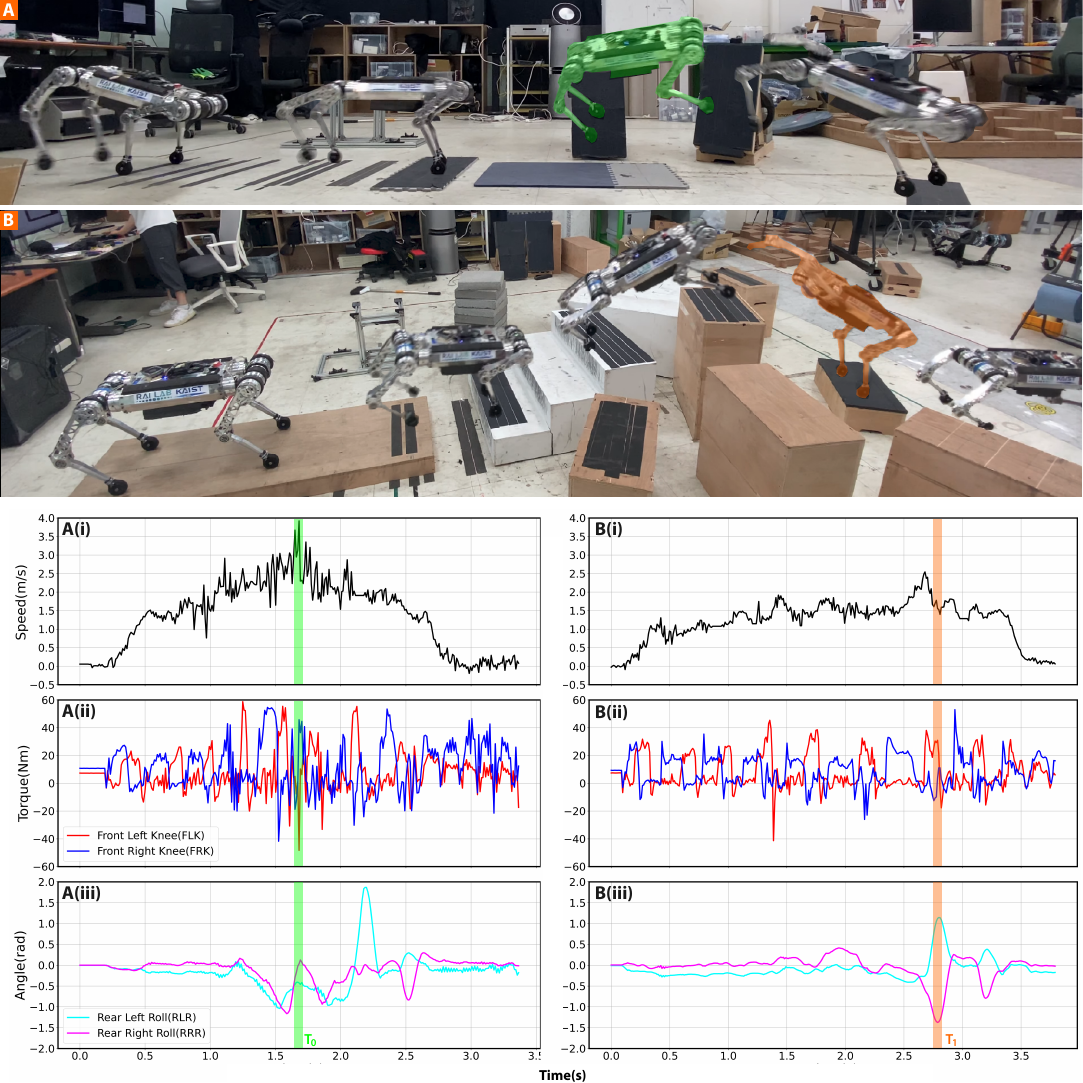}
\caption{\textbf{Real experiment analysis.} (\textbf{A},\textbf{B}) Time lapse of real experiments with the Double wall and Stepscape\_right scenarios. A(i-iii) and B(i-iii) show the robot's speed, joint torques, and angles during the experiments (\textbf{A}) and (\textbf{B}), respectively.}
\label{fig:analysis}
\end{figure}

We verified that the tracker module performed well in real-world scenarios according to the predetermined target foothold plan (Fig.~\ref{fig:scenarios}, A to G). The analysis for the \textbf{Double wall} is shown in Fig.~\ref{fig:analysis}(A and A(i) to A(iii)).
To run on double consecutive walls, the robot initially accelerated on flat ground, reaching approximately 3.0 m/s, and gained additional propulsion from the first wall, reaching speeds of around 4 m/s at time T0, as indicated in Fig.~\ref{fig:analysis}A(i).
At the moment when the robot placed its front foot on the first wall, the front left knee and front right knee joints used a high torque of about 60 Nm, which is the maximum torque of the actuator. This provided the necessary propulsion toward the next wall.

To analyze the robustness of the tracker module, real experiments were conducted with five repetitions for each of the various potential failure factors; see Fig.~\ref{fig:tracker}B and movie S2.
The first experiment [Fig.~\ref{fig:tracker}B(i)] investigated the robot’s response to a foot slip. To this end, we abraded the robot's rubber feet with sandpaper to induce slipping. From the video frames, the slip distance was estimated to be 13 cm. However, the robot still managed to land on the next desired foothold safely in five out of five trials because of the robustness of the tracker.
The second experiment [Fig.~\ref{fig:tracker}B(ii)] investigated the robot’s performance on unstable footholds. We deliberately removed the fixture pins of the box to make it movable. During the take-off phase, the box tilted by about 12°, but the tracker still managed to land successfully in five out of five trials on the next desired footholds.
The third experiment [Fig.~\ref{fig:tracker}B(iii)] evaluated the robot’s performance under map height error. To this end, one of the stepping stones was placed 15 cm higher than the map given to the planner. The robot successfully landed on the stepping stone and completed the course in five out of five trials. This suggests that the tracker is robust against mapping errors.
To quantitatively analyze the robot’s performance under mapping errors, we varied the magnitude of the map height error in simulation and measured the success rate of stable landings on the next target foothold. The success rate shown in Fig.~\ref{fig:tracker}C was measured from 300 rollouts for each map height error in the environment, with the highest difficulty level the tracker was trained on. The graph reveals that the tracker was more vulnerable when the target was perceived to be lower than it was. However, the success rate remained above 90\% even with map height errors ranging from -5 to 12 cm, demonstrating the controller's robustness to mapping error.

We also analyzed the accuracy of the tracking module across 12 scenarios depicted in Fig.~\ref{fig:scenarios}( ascent descent to flipturn ), and the results are shown in Fig.~\ref{fig:tracker}D. In the analysis, the error vector is defined as the vector from the foot’s touchdown position to the foothold target, and the tracking error is its ground projection magnitude.
On average, the simulation exhibited an error of approximately 3.7 cm, whereas the real experiment showed a slightly higher average error of 5.3 cm.
In real-world experiments, notably larger tracking errors were observed after crossing a 1.3-m gap in the 1.3-m gap scenario and after stepping on inclined surfaces in the double ramp scenario. In the 1.3-m gap scenario, the large flight phase resulting from the jump caused the state estimation error to accumulate for a longer period, leading to an increased error.

\subsubsection*{Evaluation of planner}
The planner module consisted of a sample-based optimization strategy, which involved sequential filtering to ensure that unfeasible foothold samples were quickly rejected. The first stage of sequential filtering was the performance filter, which checked whether the sampled foothold and its surrounding area were within the training range of the tracker by a carefully tuned box constraint. Subsequently, the spike filter checked the roughness of the area near the sampled foothold. Lastly, the collision filter quickly rejected the foothold sample that could potentially cause undesired collisions. It determined collision potential by comparing the robot's lowest collision boundary, estimated by the boundary estimator network, with the terrain height. 
The boundary estimator network, structured as a single multilayer perceptron (MLP), was trained via supervised learning. It demonstrated sufficient performance, with an root mean square error of 2.27 cm and an error within 3.34 cm with 95\% probability.
This boundary estimator allowed for the selection of high-quality foothold samples that account for terrain collisions without the need for rollouts, notably reducing the time complexity of planning.

\begin{figure}[p]
\includegraphics[width=\linewidth]{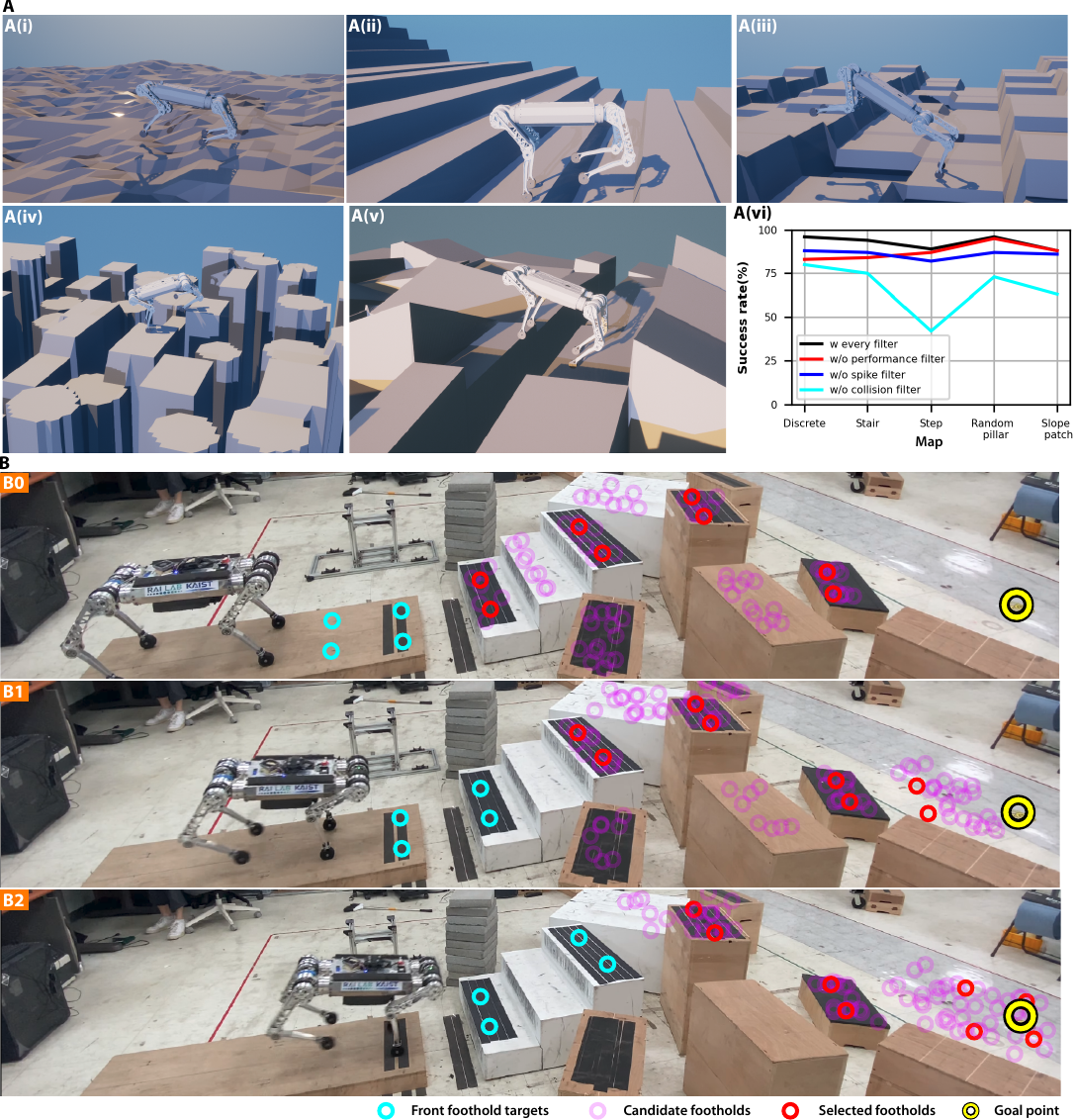}
\caption{\textbf{Evaluation of planner.} (\textbf{A}) Scenarios used for the ablation study of each filter: (\textbf{A(i)}) Discrete, (\textbf{A(ii)}) Stair, (\textbf{A(iii)}) Step, (\textbf{A(iv)}) Random pillar, and (\textbf{A(v)}) Slope patch. (\textbf{A(vi)}) Navigation success rates from the ablation study on scenarios (\textbf{A(i)} to \textbf{A(v)}). (\textbf{B}) involves the candidate foothold plans generated by the planner, the selected foothold plan with the lowest cost, the recent and next front target foothold, and the position of the goal, all shown at each timestamp during the experiment conducted in the Stepscape right scenario.}
\label{fig:planner}
\end{figure}

To analyze the importance of each filter, we conducted an ablation study on each filter using the five maps shown in Fig.~\ref{fig:planner}A(i to v). 
The names of each map are provided in the figure legend, and in the following text, each map in Fig.~\ref{fig:planner} will be referred to by its name from the legend.
The following content provides details for each map: discrete, discrete height map with height variations from $-10$ to 10 cm;
stair, stair with depth ranging from 20 to 50 cm and rise ranging from 17 to 34 cm; step, steps with widths ranging from 50 to 70 cm and heights ranging from 20 to 40 cm; random pillar, randomly placed rectangular pillars (x size, 0.3 to $\sim$ 0.4 m; and y size, 0.4 to $\sim$ 0.6 m) or cylindrical pillars (radius, 0.25 to $\sim$ 0.35 m; and height, -0.35 to $\sim$ 0.35 m); slope patch, 1.3 m–by–1.3 m slope patches tilted at random angles up to 30°.

The goal position was set to 7 to $\sim$ 8.5 m, -3 to $\sim$ 3 m from the robot's initial position, and success was defined as the robot reaching the goal position within 30 cm without undesired collisions. Success rates were measured over 500 trials for each map, and the results are shown in Fig.~\ref{fig:planner}A(vi).
The results indicated that when all filters were in place, the success rate was consistently high, with a minimum of 90\% on every map. Additionally, while all filters play important roles, the collision filter was particularly crucial in environments with a high probability of robot-terrain collisions during the flight phase, such as step and slope patch.

For brevity, we refer to the index of the target in the target buffer (Fig.~\ref{fig:pipeline}) as the target index. The target index for the front feet and the rear feet was defined as the front\_target index and the back\_target index, respectively.
To achieve successful real-world applications of the proposed hierarchical pipeline, the high-level planner must generate a feasible foothold plan before the front\_target index is updated to the next one. 
Here, the time until the front\_target index was updated was referred to as $T_{update}$. Table~\ref{tab:planning_time} displays $T_{update}$, as well as the time taken for each element of the planning process.
The first column represents the scenarios used for the time measurements. The other columns include the time taken for each filter, the time taken to rollout candidate foothold plans, the total planning time $T_{update}$, and the safety factor. The safety factor was calculated by $T_{update}$ dividing the total time consumed in the planning process, indicating how much faster the planner operated relative to the real-time threshold.
The recorded times were based on the maximum values of 8 different candidate foothold plans.
Table~\ref{tab:planning_time} includes time measurements from stepscape\_left, stepscape\_right, ringnest\_orbit, ringnest\_corecut, and flipturn scenarios from Fig.~\ref{fig:scenarios}, and discrete, stair, step, randompillar, and slope patch scenarios from Fig.~\ref{fig:planner}A. 
The time data for scenarios from Fig.~\ref{fig:scenarios} were measured during real experiments. For scenarios from Fig.~\ref{fig:planner}A, the time data for each map were the average values from 50 trials with randomly placed goal positions within a specified range in simulation. We conducted all experiments in simulation using a personal computer with the same specifications as Raibo's onboard computer.

The minimum safety factor across all scenarios was 6.35, whereas the average was 18.41. This indicated that, across various complex scenarios, the planner module computed a foothold plan at least 6.35 times faster, and on average 18.41 times faster, than the time required by the tracker to request the next foothold plan. These values provided quantitative proof that our planner was sufficiently fast for real-time planning.
The planning time was relatively short for \textbf{Discrete}, which was relatively flat and smooth. However, for very sparse maps, especially \textbf{Flipturn}, the performance filter took longer because sampled footholds were mostly rejected.
For the spike filter time, scenarios such as \textbf{Stepscape\_left}, \textbf{Stepscape\_right}, and \textbf{Stair}, which had stair-like structures with significant elevation changes between adjacent points, took a long time.
Regarding the collision filter, scenarios with fewer protrusions between foothold targets, like \textbf{Discrete} and \textbf{Random pillar}, consumed less time. In contrast, scenarios such as \textbf{Stepscape\_left}, \textbf{Stepscape\_right}, \textbf{Ringnest\_corecut}, and \textbf{Stair}, where there was a higher probability of undesired collisions, resulted in longer processing times.

Real-world experiments were successfully conducted using the planner in various complex scenarios in Fig.~\ref{fig:scenarios}( stepscape\_left to flipturn). The analysis for scenario \textbf{Stepscape\_right} is shown in Fig.~\ref{fig:analysis}B(i to iii).
In stepscape\_right , the robot navigated a complex map with stairs and obstacles of varying heights at speeds of up to 2.54 m/s. The maximum torque observed was around 52 Nm, which occurred when the robot stepped down from a high obstacle. It was also observed that during this period, the robot spread the rear left roll and rear right roll joints outward to prevent the rear legs from colliding with the obstacle. Movie 1 includes the entire trajectory of this experiment.

The candidate foothold plans and the selected foothold plan over time in stepscape\_right are shown in Fig.~\ref{fig:planner}B.
Here, Fig.~\ref{fig:planner}(B0 to B2) sequentially illustrates the moments when the robot's front\_target index changed from 0 to 1, 1 to 2, and 2 to 3, respectively.
From the time between Fig.~\ref{fig:planner}(B0) and Fig.~\ref{fig:planner}(B1), as the robot's front foot moved toward the 1st target, the tracker received the positions of the first and second foothold targets as observations for the front foot, shown as sky-blue circles. During this time, the planning module identified the eight feasible candidate foothold plans, depicted as light pink circles. Subsequently, the best foothold plan with the lowest cost, indicated by red circles, is selected through rollout in simulation. 
Movie 1 shows that the robot successfully reached the goal point (yellow circle) by tracking the foothold plan selected through this process.

\section*{Discussion}
This study has presented a hierarchical navigation pipeline that combines a time-efficient sampling-based planner with a robust learning-based tracker. It enables automatic discovery and control of fast and agile motions on complex and discrete terrains, which were difficult to achieve with previous approaches.

We evaluated our pipeline in both simulations and real-world experiments. 
First, the tracker was successfully tested on predefined foothold targets in parkour-like terrains such as gaps, consecutive ramps, and walls. The average tracking error of approximately 5.3 cm in real-world maps suggests that the tracker, learned through the proposed framework, can accurately step on the target footholds.
The generative model-based map generator created an environment with the desired difficulty on the basis of the current tracker performance, providing meaningful reward variations and facilitating stable learning. As the competitive training progressed, the range of environments provided by the map generator gradually expanded within the physically feasible limits of the robot, ultimately enabling the tracker to achieve notably high performance as demonstrated by wall-running behavior.
This approach is not limited to just our task and can be applied to other practices where delicate environmental planning is required to develop high-performance controllers.

Next, the planner was tested on terrains composed of various sizes of stepping stones, stairs, and ramps, and these results support the planner’s efficiency in complex and challenging environments. Because the planner is independent of map distribution, it consistently provided safe and feasible foothold plans on diverse and complex terrains. Given the time-efficient filtering structure of the planner, feasible foothold plans can be computed using the onboard computer in real time with high safety factors. The rollout process in physics simulation thoroughly considers the robot's full dynamics, thereby ensuring the stability of the selected foothold plan. Furthermore, by adjusting the cost function used in this process, the user can guide the robot's movement according to desired objectives.

\subsubsection*{Limitations and future work}
In this study, the planner module searched for footholds on a 2.5-dimensional (2.5D) map. Consequently, despite the tracker module's ability to perform multiple wall runs, the planner module cannot provide foothold plans for vertical walls. Future work will make our planner compatible with 3D map representation, such as a 3D voxel map, to generate a foothold plan for vertical wall running as well.
In addition, the current use of a motion capture system prevented experiments from being conducted in outdoor environments. In the future, we will combine our pipeline with a perception system capable of predicting occluded regions and robust to high-speed acceleration to enable autonomous navigation in outdoor settings.

Although further research is needed to address the aforementioned limitations, this study presents a learning framework for training controllers capable of high-speed navigation on discrete terrains, along with a planning method that provides feasible foothold plans in real time.

\section*{Materials and methods}
\subsubsection*{Overview}
This research aimed to develop a navigation system that can quickly and safely reach a goal position across complex and discontinuous terrain. We decomposed this challenging problem into two parts: finding a feasible foothold plan (planner) and accurately tracking this foothold plan (tracker).
In particular, we drastically reduced the time complexity of planning through two settings: using only the foothold positions as the search space, unlike previous studies~\cite{mastalli2015line,risbourg2022real} that considered additional factors such as contact phase and base pose, and ensuring that rear feet step where the front feet have already stepped.
To develop the pipeline, we first trained the tracker module to maximize the robot's tracking performance. During this phase, the foothold target distribution was provided by a generative model called a map generator. It was trained concurrently and competitively with the tracker, ensuring that the tracker was trained at a desired difficulty level. Subsequently, we designed the sampling-based planner module that can generate a plan that reflects the capabilities and characteristics of the trained tracker module. Details of all of the networks comprising the pipeline are provided in the Supplementary Methods "Network details."

\subsubsection*{Training procedure}
The tracker module was trained via reinforcement learning with the proximal policy optimization \cite{schulman2017proximal} algorithm, with its parameters in table~\ref{tab:ppo_parameters}.
We generated 300 environments in parallel using the Raisim \cite{raisim} physics simulation, with 4.2-second episodes, a 2 ms simulation time step, and a 0.01 s control time step.
Each environment consisted of 10 stepping stones placed consecutively, as shown in Fig.~\ref{fig: mapgenerator}D. The process for generating these stepping stones is detailed in the section \textbf{Terrain generation}.

The tracker was trained to accurately step onto the target located on the stepping stone corresponding to the target index. At this time, the left and right feet shared the same target index, resulting in simultaneous target updates.
The condition for the target updater to update the front\_target index or back\_target index was when either the left or right foot has been in contact with the target for a specified time (0.06 s).

To stabilize and expedite the training process, we implemented two termination conditions: (i) if any internal contact occurs within the robot or if any part of the robot, other than the feet, touches the terrain, and (ii) if the normal vector of the robot's base tilts more than 110 degrees.

Furthermore, to reduce the sim-to-real gap and enhance the robustness of the controller, we applied 10 types of domain randomization, such as PD gain and observation noise, during the training stage. Details are provided in table~\ref{tab:domain_randomization}.

\subsubsection*{Terrain generation}
In the training environment, the pose of each stepping stone relative to the previous one is represented by a 6D parameter called $\psi$.
The schematic diagram of the components that make up $\psi$ is shown in Fig.~\ref{fig: mapgenerator}A, with detailed explanations provided in Supplementary Methods "Components of psi."

The training has two stages. In the initial stage, the range of $\psi$ is increased at a fixed rate as described in table~\ref{tab:linear_curricula}. This stage is used for generating sufficient data to initially train the map generator and to stabilize the learning process. 
In the second stage, the map generator evolves competitively with the tracker, providing $\psi$, which generates environments with desired difficulty.

\begin{figure}[p]
\includegraphics[width=\linewidth]{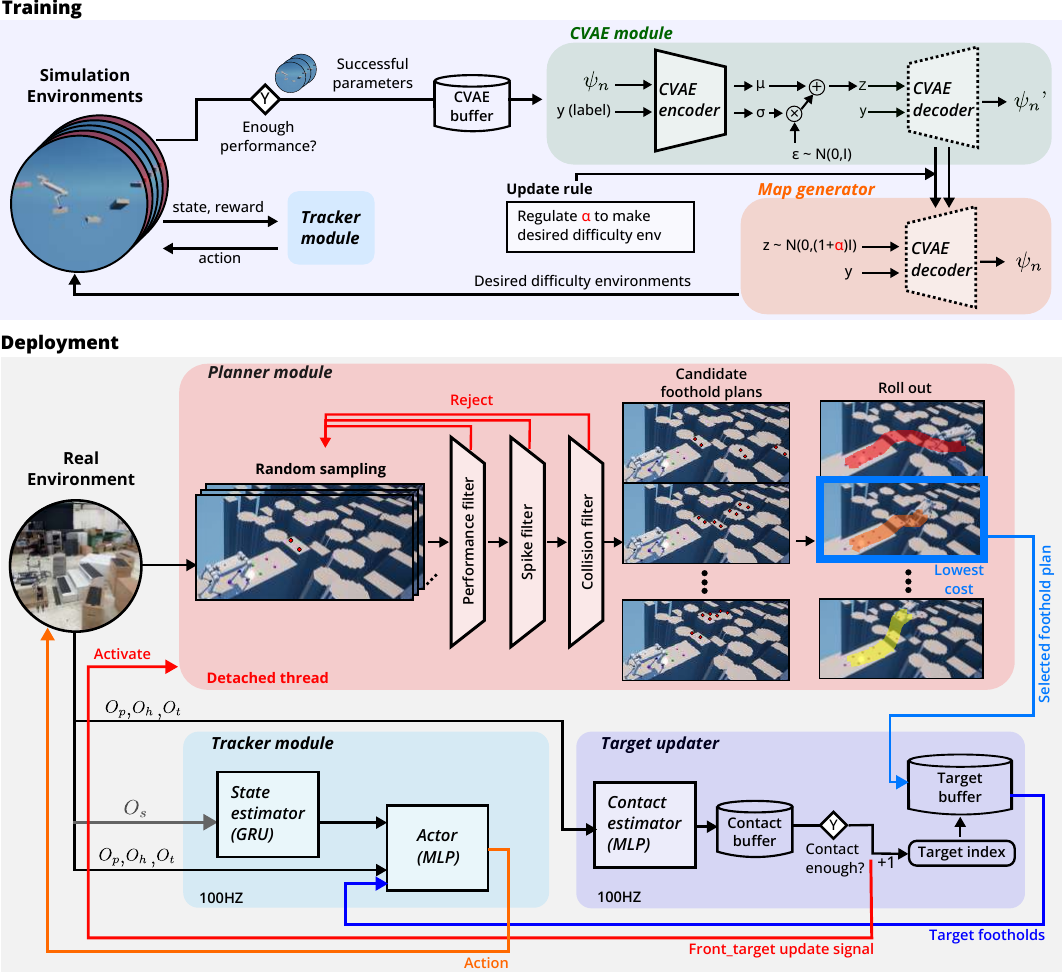}
\caption{\textbf{Overall pipeline.} 
In \textbf{Training stage}, the tracker module is competitively trained with a map generator. Whenever the tracker achieves sufficient performance, the CVAE module is retrained using the successful map parameters, and the decoder part is used as the map generator. By adjusting the alpha value, the map generator provides the desired difficulty map to the agent.
In \textbf{Deployment stage}, when the robot steps on the target for a sufficient time, the target index is updated. Whenever the front foot's target index is updated, the planner module is activated on the detached thread to generate a feasible foothold plan. The planner module generates candidate foothold plans using a sampling-based strategy with a consecutive filtering structure, rapidly rejecting risky and unfeasible samples. Candidate plans are evaluated through the rollout process in physics simulation, and the plan with the lowest cost is selected.}
\label{fig:pipeline}
\end{figure}

We used a generative model, conditional variational autoencoder(CVAE)\cite{CVAE}, to represent the distribution of $\psi$. The structure of this model is shown in Fig.~\ref{fig:pipeline}, and the map generator refers only to the decoder part of the CVAE (Fig.~\ref{fig: mapgenerator}B).
Both the CVAE encoder and decoder are of an MLP \cite{mlp}. Here, the label y includes the two previous $\psi$s and $T_{last}$, which indicates whether the next target is the last one.
For competitive learning with the tracker, we retrain the CVAE module to provide more challenging environments whenever the tracker achieves sufficient performance. The criterion for sufficient performance is whether the average number of stepping stones overcome exceeds 9.3 out of 10 stepping stones per episode.
The data for the CVAE training are the set of $\psi$ and $T_{last}$ that the tracker has successfully overcome, and the loss function consists of reconstruction loss and KL-divergence loss. Details for CVAE are provided in Supplementary Methods "Network details."
After retraining the CVAE module, the difficulty of the maps generated by the map generator was adjusted using $\alpha$. This value regulates the variance of the distribution from which the latent vector $Z_{latent}$ is sampled when generating $\psi$ parameters. 
The detailed algorithm for this adversarial training process is represented in algorithm S1.

The distribution of $\psi$ from the map generator, at different stages of training, is shown in Fig.~\ref{fig: mapgenerator}C, and examples of the resulting environments are in Fig.~\ref{fig: mapgenerator}D.
In Fig.~\ref{fig: mapgenerator}C, the first row comprises $r$-$\phi$ graphs. The graphs show that, as the number of updates increases to 30, the range of $r$ becomes wide, with a minimum of 0.4 m to a maximum of 1.6 m. When the value of $r$ is small, the range of generated $\phi$ is small because a large positive $\phi$ increases the likelihood of colliding with the bottom of the next stepping stone during the flight phase, while a large negative $\phi$ increases the risk of hitting the top of the current stepping stone.
On the contrary, as the $r$ increases, a wide range of $\phi$ from $-60^\circ$ to $60^\circ$ is generated.

The graphs in the second row are $\theta$-$x\_tilt$ graph. These graphs show a strong correlation between the two parameter distributions. Physically, this indicates that the surface normal vector of the next stepping stone is oriented opposite to the lateral direction of the jump so that it can allow greater deceleration of the robot.
In the final stage, after 30 updates, the map generator allows the tracker to experience wall-running maps with an $x\_tilt$ of 90 degrees. Additionally, it generates maps including stepping stones with a distance of 1.6 m and a $60^\circ$ inclination angle of steps in the simulation. In such environments, the tracker successfully reaches the last target at a success rate of over 95\%. We can observe that the proposed map generator effectively captures the region of physically feasible parameters during the evolutionary process and enables the tracker to overcome highly challenging terrains.

In addition, we validated that the proposed map generator method is more effective in environments characterized by high-dimensional parameters compared with existing curriculum methods. The details of this validation can be found in the Supplementary Results section "Analysis of various curricula."

\subsubsection*{Input representation for Actor and State Estimator}
The tracker module is composed of an actor with an MLP structure and a state estimator network with a gated recurrent unit (GRU) \cite{gru}.
As in \cite{ji2022concurrent}, the module concatenates the output of the state estimator network with the observation to provide input to the actor.
The input to the actor, denoted as $O_{a}$, is composed of $O_p$,$O_h$, $O_t$, and $unObs$.
$O_p$ represents the robot's proprioceptive state, consisting of body orientation, body angular velocity, joint positions, and joint velocities.
$O_h$ is the history observation that includes previous joint positions, joint velocities, and joint targets.
$O_t$ consists of target-related information. This includes the vector from the center of mass (COM) of the robot's base to foothold targets in the foothold target plan in the robot frame at the moment the target index is updated. Here, the foothold target plan includes the next $num_{target}$ targets for each foot. If $num_{target}$ is too small, the robot might struggle to reach future targets, while a large value increases time complexity in planning. Simulations show that 2 is a practical value, balancing performance and time complexity.
$O_t$ also includes $time_{front}$ and $time_{back}$, the elapsed time since the front\_target index and back\_target index updated, respectively.
$unObs$ is provided by the state estimator network, representing the robot's linear velocity in the robot frame.
The input to the state estimator, denoted as $O_s$, consists of $O_p$ and $prev\_action$. $prev\_action$ refers to the action taken 0.01 s earlier.

\subsubsection*{Reward function for Actor}
The rewards are broadly categorized into target-related, constraint-related, and style-related rewards. The coefficients and formulas for each reward term are provided in table~\ref{tab:reward_function}.
Even with only six rewards (target, bound, joint velocity, and torque), the tracker can be trained successfully in simulation. The remaining rewards were added for better real-world performance and to produce more natural motions.
The following explains the target-related rewards.
The target\_sparse reward encourages the foot position at touchdown to be close to the foothold target.
The target\_dense reward acts as an auxiliary reward that increases as the foot moves closer to the next target. 
The target\_last reward is given when the robot reaches the last target.

Constraint-related reward includes only one reward, joint\_limit\_reward. It ensures joints do not exceed their limits.

Style-related rewards guide the robot to achieve the desired movements and ensure robust control in real-world scenarios.
The following is a description of the components. Torque reward is for minimizing torque usage. Slip reward penalizes foot velocity parallel to the stepping stone during the stance phase. Foot\_gather\_lateral reward is to encourage gathering the left and right feet to a distance of 25 cm when touching down the terrain. Foot\_gather\_longitudinal reward is to encourage gathering the front foot and rear foot when the front foot lifts off. Bound reward is to promote a bounding gait. Joint velocity reward is to minimize joint velocity. Stop reward is to minimize the linear and angular velocity of the robot's base when reaching the last target. Impact reward is to reduce the linear velocity of the foot perpendicular to the stepping stone's upper plane at touchdown. Smooth reward is to prevent oscillatory action.

During the training stage, the target\_last\_reward encourages the controller to quickly place the foot on the stepping stone, thereby quickly updating the target index and facilitating faster progress toward the final target. The increase in touchdown speed caused the foot to bounce upon touchdown in real-world experiments, leading to control instability and an increase in tracking error. The impact reward mitigates this issue by reducing the touchdown speed (see movie S3).

\subsubsection*{Planner module}
The planner module operates on detached threads to generate a feasible foothold plan each time the front\_target index is updated. The process begins with iterative sampling and filtering, resulting in eight candidate foothold plans. These plans are then evaluated in a physics simulation to ensure their physical consistency and performance. The best foothold plan is selected on the basis of these simulation evaluations. In the following paragraphs, we detail each stage of this planning process, with further information provided in Supplementary Methods "Details of sampling and filtering footholds."

The first stage is sampling with sequential filtering. This stage begins with the random sampling of potential footholds. Given the vast sampling space for foothold plans, we used two strategies to narrow the search space: synchronizing the update timing of the targets for the two front feet and two rear feet, and having the rear foot step where the corresponding front foot previously stepped. These simplifications allow us to sample only one foothold per gait cycle. 
To sample a foothold, we randomly picked $r$, $\theta$, and $\Delta_{yaw}$ considering the tracker’s final training range.
This allows us to consider the robot's dynamics, unlike previous research \cite{mastalli2015line, risbourg2022real,deits2014footstep} that only considers the kinematics for foothold feasibility.

\begin{figure}[p]
\includegraphics[width=\linewidth]{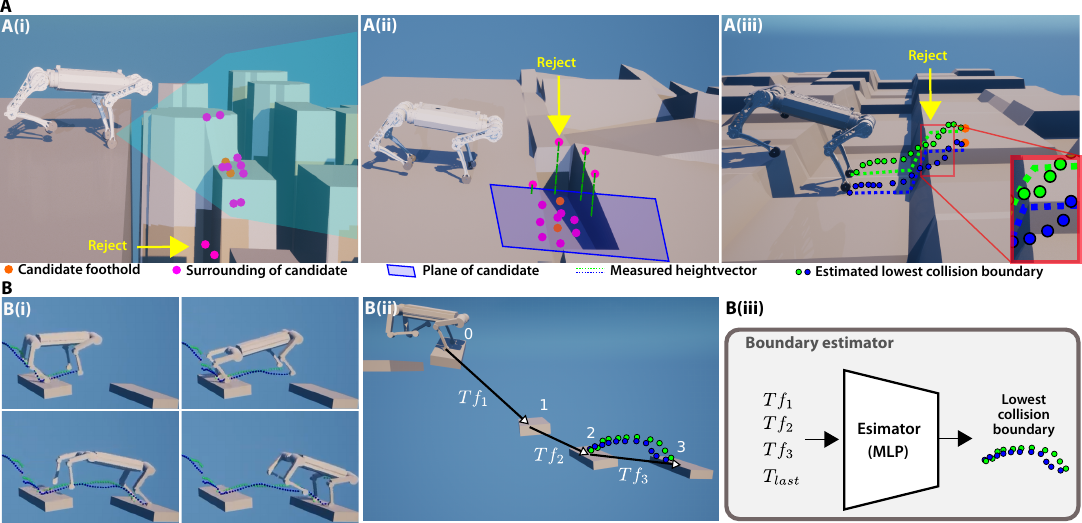}
\caption{\textbf{Structure of filters.} \textbf{(A)} Rejected cases by each filter: (i) performance filter, (ii) spike filter, and (iii) collision filter. \textbf{(B)} (i) change in the lowest collision boundary as the robot moves forward. (ii) Vectors Tf1, Tf2, and Tf3 provided as boundary estimator network input. (iii) Structure of the boundary estimator network.}
\label{fig:filters}
\end{figure}

Each generated sample underwent sequential filtering to quickly reject unfeasible footholds. A diagram of the filtering structures is shown in Fig.~\ref{fig:filters}A. The first filter in this stage is the performance filter, which verifies whether the heights of the sampled foothold and its surroundings fall within the tracker’s training range by a carefully tuned box constraint.

The second filter is the spike filter, which rejects footholds if the surrounding regions exhibit high variance in height. It uses principal components analysis to calculate the slope and assess the deviation from a linear model. This filter ensures that the foothold regions are free from excessive curvature or roughness.

Lastly, the collision filter compares the lowest collision boundary of the robot, predicted by the boundary estimator network, with the height of the map to identify potential collisions along the robot’s motion toward the next target.
The boundary estimator is a single MLP as shown in Fig.~\ref{fig:filters}B(iii), and it is trained through supervised learning to predict the collision potential. The input $Tf_k$ represents the vector from the (k-1)-th target to the k-th target (Fig.~\ref{fig:filters}B(ii)), and $T_{last}$ is a variable indicating whether the sampled foothold is the last target where the robot should stop.
Note that what it predicts is not simply the trajectory of a single point on the robot's collision body(Fig.~\ref{fig:filters}B(i)). Instead, it predicts the lowest point on the shape formed by sweeping the collision bodies along the trajectory. This point does not correspond to any specific location on the robot.
The process of training this network is detailed in the Supplementary Methods "Data collection process of boundary estimator."

We repeated the process of sampling and filtering to form each foothold pair one by one until we obtained eight foothold plans that are four-foothold-pair-long. Parallelization on the CPU enables fast computation of this process.

All plans are evaluated in a physics engine to select the best foothold plan. 
They are simulated with the trained tracker and the complete model of the Raibo robot in eight separate threads. Through this process, invalid plans are rejected, and the cost associated with each candidate foothold plan is calculated.
The cost function in this process can be designed on the basis of the user's preferences. The following outlines the cost functions used in our experiment.

Survive cost refers to the cost associated with termination during the rollout process. The distance cost measures how close the robot and the goal position are after the rollout. The direction cost accounts for the angle between the robot's displacement vector during the rollout and the vector from the initial position to the goal. Lastly, the elevation cost evaluates the cost based on how much the robot ascends and descends throughout the rollout. Detailed information is provided in table~\ref{tab:planning_cost}. The overall process of the planner module is presented in algorithm S2.

\subsubsection*{Statistical analysis}
\begin{figure}[p]
\includegraphics[width=\linewidth]{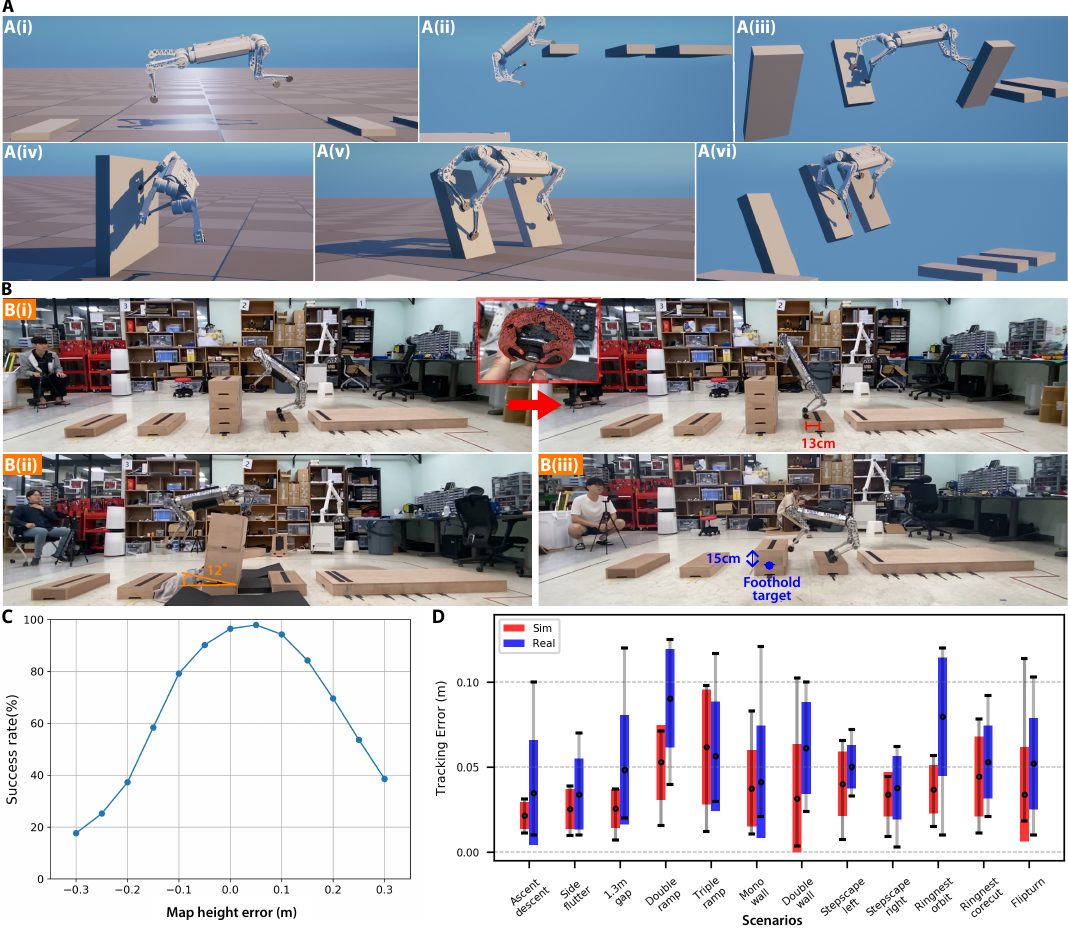}
\caption{\textbf{Evaluation of tracker.} \textbf{(A)} Various extreme parkour behaviors in simulation. (\textbf{A(i)}) 1.9 m gap (4x leg length), (\textbf{A(ii)}) 1 m vertical jump (2.1x leg length), (\textbf{A(iii)}) 70° triple ramp, (\textbf{A(iv)}) mono wall, (\textbf{A(v)}) double wall, (\textbf{A(vi)}) triple wall.
(\textbf{B}) Measuring robustness against various failure factors :
(\textbf{B(i)}) foot slippage, (\textbf{B(ii)}) unstable footholds, and (\textbf{B(iii)}) map height errors. (\textbf{C}) Success rates across various map height errors in simulation. (\textbf{D}) Tracking error measured in simulation and the real world across 12 scenarios in Fig.~\ref{fig:scenarios} (sample size N=10). Each colored bar represents the interval corresponding to the $\text{mean} \pm \text{SD}$. The black circle denotes the sample mean, while the black whiskers indicate the maximum and minimum values observed, respectively.}
\label{fig:tracker}
\end{figure}

The mean and the SD shown in Fig.~\ref{fig:tracker}D for sample size N were calculated using the following equation:
\begin{equation}
\mu = \frac{1}{N} \sum_{i=1}^{N} x_i
\quad \text{,} \quad
\sigma = \sqrt{\frac{1}{N} \sum_{i=1}^{N} (x_i - \mu)^2}
\end{equation}
Here, $\mu$ is the mean and $\sigma$ is the SD.


\begin{figure}[p]
\includegraphics[width=\linewidth]{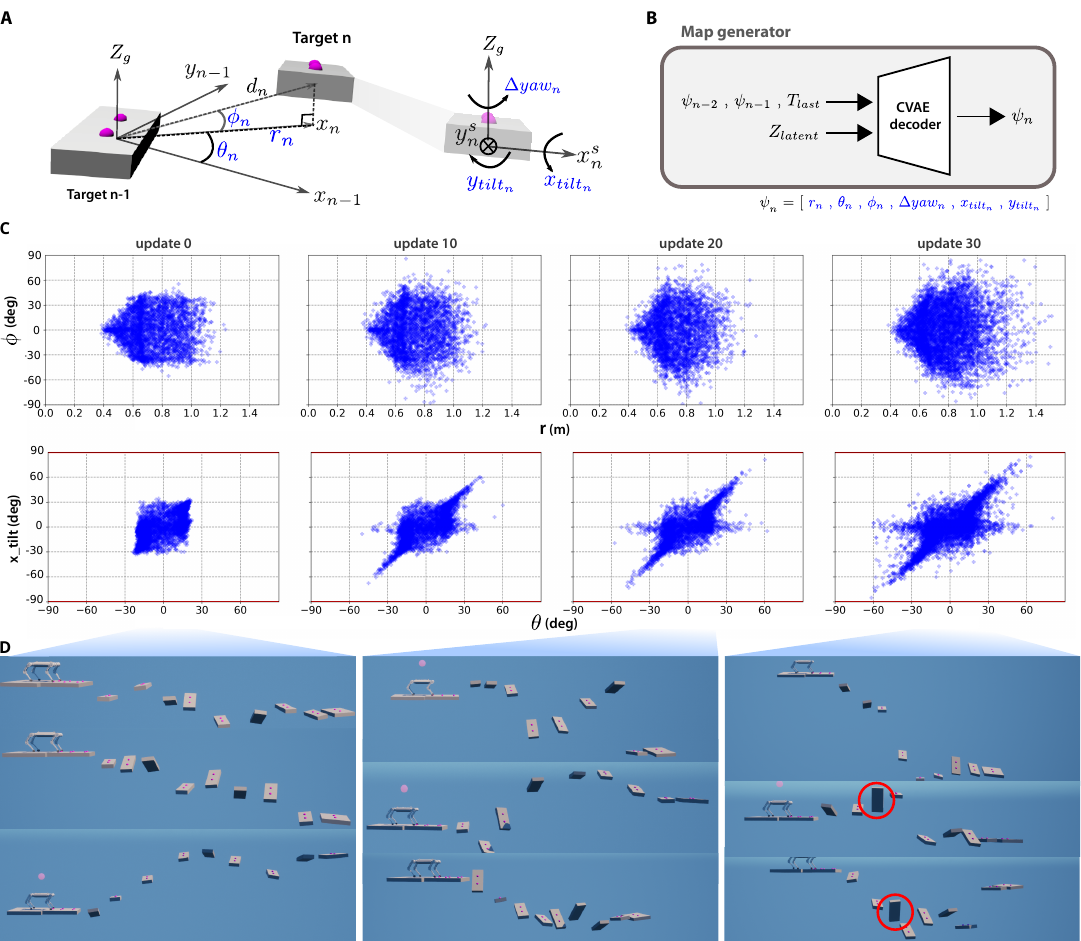}
\caption{\textbf{Evolution of map generator.} (\textbf{A}) Diagram of the components of the $\psi$ parameter that constitutes a pose of stepping stone. $d_n$ represents the vector from the center of the (n-1)th stepping stone to the center of the nth stepping stone. The vector $x_n$ is the projection of $d_n$ onto the ground plane. The vector $Z_g$ is the vector in the direction opposite to gravity. $x_n^s$ and $y_n^s$ represent the $x$ and $y$ axes of the nth stepping stone, respectively. (\textbf{B}) The structure of the map generator. (\textbf{C}) $\psi$ distribution based on the number of updates to the map generator. (\textbf{D}) Examples of generated environments. Red circles indicate regions where stepping stones are nearly vertical to the ground.}
\label{fig: mapgenerator}
\end{figure}

\newpage
\begin{table}[ht]
    \centering
    \small
    \caption{\textbf{Planning time analysis on various scenarios}}
    \label{tab:planning_time}
    \begin{tabular}{|c|c|c|c|c|c|c|c|}
        \hline
        \textbf{Scenario} & \shortstack[c]{\textbf{Performance}\\\textbf{filter(s)}} & \shortstack[c]{\textbf{Spike}\\\textbf{filter(s)}} & \shortstack[c]{\textbf{Collision}\\\textbf{filter(s)}} & \shortstack[c]{\textbf{Roll out(s)}} & \shortstack[c]{\textbf{Total}\\\textbf{planning (s)}} & \shortstack[c]{\textbf{$T_{update}$ (s)}} & \shortstack[c]{\textbf{Safety}\\\textbf{factor}} \\
        \noalign{\hrule height 3pt}
        \textbf{Stepscape\_left} & 0.002242 & 0.007671 & 0.010655 & 0.034563 & 0.055131 & 0.365000 & 6.620622 \\
        \hline
        \textbf{Stepscape\_right} & 0.002847 & 0.007214 & 0.012852 & 0.035674 & 0.058587 & 0.372000 & 6.349528 \\
        \hline
        \textbf{Ringnest\_orbit} & 0.003044 & 0.000799 & 0.001790 & 0.007806 & 0.013438 & 0.408571 & 30.404265 \\
        \hline
        \textbf{Ringnest\_corecut} & 0.009594 & 0.001445 & 0.021793 & 0.021980 & 0.054812 & 0.398000 & 7.261226 \\
        \hline
        \textbf{Flipturn} & 0.015484 & 0.001764 & 0.007821 & 0.012123 & 0.037193 & 0.408750 & 10.990092 \\
        \hline
        \textbf{Discrete} & 0.000000 & 0.000107 & 0.000231 & 0.011451 & 0.011790 & 0.365000 & 30.959237 \\
        \hline
        \textbf{Stair} & 0.000035 & 0.003683 & 0.010537 & 0.021287 & 0.035542 & 0.376364 & 10.589195 \\
        \hline
        \textbf{Step} & 0.000017 & 0.001606 & 0.001697 & 0.012514 & 0.015835 & 0.386000 & 24.376985 \\
        \hline
        \textbf{Random pillar} & 0.000646 & 0.001451 & 0.002921 & 0.009830 & 0.014849 & 0.376667 & 25.367239 \\
        \hline
        \textbf{Slope patch} & 0.000209 & 0.001292 & 0.002539 & 0.010148 & 0.014187 & 0.443000 & 31.225912 \\
        \hline
    \end{tabular}
\end{table}

\clearpage
\bibliography{discrete_navigation} 

\begin{thebibliography}{10}
\providecommand{\url}[1]{\texttt{#1}}
\expandafter\ifx\csname urlstyle\endcsname\relax
  \providecommand{\doi}[1]{doi:\discretionary{}{}{}#1}\else
  \providecommand{\doi}{doi:\discretionary{}{}{}\begingroup \urlstyle{rm}\Url}\fi

\bibitem{katz2019mini}
B.~Katz, J.~Di~Carlo, S.~Kim, Mini cheetah: A platform for pushing the limits of dynamic quadruped control, in \emph{2019 international conference on robotics and automation (ICRA)} (IEEE) (2019), pp. 6295--6301.

\bibitem{hutter2016anymal}
M.~Hutter, \emph{et~al.}, Anymal-a highly mobile and dynamic quadrupedal robot, in \emph{2016 IEEE/RSJ international conference on intelligent robots and systems (IROS)} (IEEE) (2016), pp. 38--44.

\bibitem{shin2022design}
Y.-H. Shin, \emph{et~al.}, Design of KAIST HOUND, a quadruped robot platform for fast and efficient locomotion with mixed-integer nonlinear optimization of a gear train, in \emph{2022 International Conference on Robotics and Automation (ICRA)} (IEEE) (2022), pp. 6614--6620.

\bibitem{semini2011design}
C.~Semini, \emph{et~al.}, Design of HyQ--a hydraulically and electrically actuated quadruped robot. \emph{Proceedings of the Institution of Mechanical Engineers, Part I: Journal of Systems and Control Engineering} \textbf{225}~(6), 831--849 (2011).

\bibitem{miki2022learning}
T.~Miki, \emph{et~al.}, Learning robust perceptive locomotion for quadrupedal robots in the wild. \emph{Sci. Robot.} \textbf{7}~(62), eabk2822 (2022).

\bibitem{shi2024rethinking}
F.~Shi, \emph{et~al.}, Rethinking Robustness Assessment: Adversarial Attacks on Learning-based Quadrupedal Locomotion Controllers. \emph{arXiv:2405.12424}  (2024).

\bibitem{miki2024learning}
T.~Miki, J.~Lee, L.~Wellhausen, M.~Hutter, Learning to walk in confined spaces using 3d representation. \emph{arXiv:2403.00187}  (2024).

\bibitem{jenelten2022tamols}
F.~Jenelten, R.~Grandia, F.~Farshidian, M.~Hutter, TAMOLS: Terrain-aware motion optimization for legged systems. \emph{T-RO} \textbf{38}~(6), 3395--3413 (2022).

\bibitem{abdalla2023efficient}
A.~Abdalla, M.~Focchi, R.~Orsolino, C.~Semini, An Efficient Paradigm for Feasibility Guarantees in Legged Locomotion. \emph{T-RO} \textbf{39}~(5), 3499--3515 (2023).

\bibitem{di2018dynamic}
J.~Di~Carlo, P.~M. Wensing, B.~Katz, G.~Bledt, S.~Kim, Dynamic locomotion in the mit cheetah 3 through convex model-predictive control, in \emph{2018 IEEE/RSJ international conference on intelligent robots and systems (IROS)} (IEEE) (2018), pp. 1--9.

\bibitem{kim2019highly}
D.~Kim, J.~Di~Carlo, B.~Katz, G.~Bledt, S.~Kim, Highly dynamic quadruped locomotion via whole-body impulse control and model predictive control. \emph{arXiv:1909.06586}  (2019).

\bibitem{chen2023quadruped}
H.~Chen, Z.~Hong, S.~Yang, P.~M. Wensing, W.~Zhang, Quadruped capturability and push recovery via a switched-systems characterization of dynamic balance. \emph{IEEE T-RO} \textbf{39}~(3), 2111--2130 (2023).

\bibitem{minniti2021adaptive}
M.~V. Minniti, R.~Grandia, F.~Farshidian, M.~Hutter, Adaptive CLF-MPC with application to quadrupedal robots. \emph{IEEE Robot. Autom. Lett.} \textbf{7}~(1), 565--572 (2021).

\bibitem{li2024cafe}
H.~Li, P.~M. Wensing, Cafe-mpc: A cascaded-fidelity model predictive control framework with tuning-free whole-body control. \emph{arXiv:2403.03995}  (2024).

\bibitem{garcia2021time}
G.~Garc{\'\i}a, R.~Griffin, J.~Pratt, Time-varying model predictive control for highly dynamic motions of quadrupedal robots, in \emph{2021 IEEE International Conference on Robotics and Automation (ICRA)} (IEEE) (2021), pp. 7344--7349.

\bibitem{hong2020real}
S.~Hong, J.-H. Kim, H.-W. Park, Real-time constrained nonlinear model predictive control on so (3) for dynamic legged locomotion, in \emph{2020 IEEE/RSJ International Conference on Intelligent Robots and Systems (IROS)} (IEEE) (2020), pp. 3982--3989.

\bibitem{nguyen2022continuous}
C.~Nguyen, L.~Bao, Q.~Nguyen, Continuous jumping for legged robots on stepping stones via trajectory optimization and model predictive control, in \emph{2022 IEEE 61st Conference on Decision and Control (CDC)} (IEEE) (2022), pp. 93--99.

\bibitem{grandia2023perceptive}
R.~Grandia, F.~Jenelten, S.~Yang, F.~Farshidian, M.~Hutter, Perceptive locomotion through nonlinear model-predictive control. \emph{IEEE T-RO} \textbf{39}~(5), 3402--3421 (2023).

\bibitem{agrawal2022vision}
A.~Agrawal, S.~Chen, A.~Rai, K.~Sreenath, Vision-aided dynamic quadrupedal locomotion on discrete terrain using motion libraries, in \emph{2022 International Conference on Robotics and Automation (ICRA)} (IEEE) (2022), pp. 4708--4714.

\bibitem{choi2023learning}
S.~Choi, \emph{et~al.}, Learning quadrupedal locomotion on deformable terrain. \emph{Sci. Robot.} \textbf{8}~(74), eade2256 (2023).

\bibitem{ji2022concurrent}
G.~Ji, J.~Mun, H.~Kim, J.~Hwangbo, Concurrent training of a control policy and a state estimator for dynamic and robust legged locomotion. \emph{IEEE Robot. Autom. Lett.} \textbf{7}~(2), 4630--4637 (2022).

\bibitem{kumar2021rma}
A.~Kumar, Z.~Fu, D.~Pathak, J.~Malik, Rma: Rapid motor adaptation for legged robots. \emph{arXiv:2107.04034}  (2021).

\bibitem{feng2023genloco}
G.~Feng, \emph{et~al.}, Genloco: Generalized locomotion controllers for quadrupedal robots, in \emph{Conference on Robot Learning} (PMLR) (2023), pp. 1893--1903.

\bibitem{peng2018sim}
X.~B. Peng, M.~Andrychowicz, W.~Zaremba, P.~Abbeel, Sim-to-real transfer of robotic control with dynamics randomization, in \emph{2018 IEEE international conference on robotics and automation (ICRA)} (IEEE) (2018), pp. 3803--3810.

\bibitem{kim2024constraint}
Y.~Kim, \emph{et~al.}, Not only rewards but also constraints: Applications on legged robot locomotion. \emph{IEEE T-RO}  (2024).

\bibitem{jenelten2024dtc}
F.~Jenelten, J.~He, F.~Farshidian, M.~Hutter, DTC: Deep Tracking Control. \emph{Sci. Robot.} \textbf{9}~(86), eadh5401 (2024).

\bibitem{hoeller2024anymal}
D.~Hoeller, N.~Rudin, D.~Sako, M.~Hutter, Anymal parkour: Learning agile navigation for quadrupedal robots. \emph{Sci. Robot.} \textbf{9}~(88), eadi7566 (2024).

\bibitem{cheng2023extreme}
X.~Cheng, K.~Shi, A.~Agarwal, D.~Pathak, Extreme parkour with legged robots. \emph{arXiv:2309.14341}  (2023).

\bibitem{duan2022learning}
H.~Duan, \emph{et~al.}, Learning dynamic bipedal walking across stepping stones, in \emph{2022 IEEE/RSJ International Conference on Intelligent Robots and Systems (IROS)} (IEEE) (2022), pp. 6746--6752.

\bibitem{rudin2022advanced}
N.~Rudin, D.~Hoeller, M.~Bjelonic, M.~Hutter, Advanced skills by learning locomotion and local navigation end-to-end, in \emph{2022 IEEE/RSJ International Conference on Intelligent Robots and Systems (IROS)} (IEEE) (2022), pp. 2497--2503.

\bibitem{jeon2023learning}
S.~Jeon, M.~Jung, S.~Choi, B.~Kim, J.~Hwangbo, Learning whole-body manipulation for quadrupedal robot. \emph{IEEE Robot. Autom. Lett.} \textbf{9}~(1), 699--706 (2023).

\bibitem{wang2019}
R.~Wang, J.~Lehman, J.~Clune, K.~O. Stanley, Paired open-ended trailblazer (poet): Endlessly generating increasingly complex and diverse learning environments and their solutions. \emph{arXiv:1901.01753}  (2019).

\bibitem{lee2020}
J.~Lee, J.~Hwangbo, L.~Wellhausen, V.~Koltun, M.~Hutter, Learning quadrupedal locomotion over challenging terrain. \emph{Sci. Robot.} \textbf{5}~(47), eabc5986 (2020).

\bibitem{xie2020allsteps}
Z.~Xie, H.~Y. Ling, N.~H. Kim, M.~van~de Panne, Allsteps: curriculum-driven learning of stepping stone skills, in \emph{Computer Graphics Forum} (Wiley Online Library), vol.~39 (2020), pp. 213--224.

\bibitem{lee2024wheelnavi}
J.~Lee, \emph{et~al.}, Learning robust autonomous navigation and locomotion for wheeled-legged robots. \emph{Sci. Robot.} \textbf{9}~(89), eadi9641 (2024).

\bibitem{zhang2024resilient}
C.~Zhang, \emph{et~al.}, Resilient legged local navigation: Learning to traverse with compromised perception end-to-end, in \emph{41st IEEE Conference on Robotics and Automation (ICRA 2024)} (2024).

\bibitem{kim2022learning}
Y.~Kim, C.~Kim, J.~Hwangbo, Learning forward dynamics model and informed trajectory sampler for safe quadruped navigation. \emph{arXiv:2204.08647}  (2022).

\bibitem{chestnutt2007navigation}
J.~Chestnutt, \emph{Navigation planning for legged robots} (Carnegie Mellon University) (2007).

\bibitem{wellhausen2021rough}
L.~Wellhausen, M.~Hutter, Rough terrain navigation for legged robots using reachability planning and template learning, in \emph{2021 IEEE/RSJ International Conference on Intelligent Robots and Systems (IROS)} (IEEE) (2021), pp. 6914--6921.

\bibitem{xu2021contact}
P.~Xu, \emph{et~al.}, Contact sequence planning for hexapod robots in sparse foothold environment based on Monte-Carlo tree. \emph{IEEE Robot. Autom. Lett.} \textbf{7}~(2), 826--833 (2021).

\bibitem{tsounis2020deepgait}
V.~Tsounis, M.~Alge, J.~Lee, F.~Farshidian, M.~Hutter, Deepgait: Planning and control of quadrupedal gaits using deep reinforcement learning. \emph{IEEE Robot. Autom. Lett.} \textbf{5}~(2), 3699--3706 (2020).

\bibitem{heess2017emergence}
N.~Heess, \emph{et~al.}, Emergence of locomotion behaviours in rich environments. \emph{arXiv:1707.02286}  (2017).

\bibitem{caluwaerts2023barkour}
K.~Caluwaerts, \emph{et~al.}, Barkour: Benchmarking animal-level agility with quadruped robots. \emph{arXiv:2305.14654}  (2023).

\bibitem{peng2017deeploco}
X.~B. Peng, G.~Berseth, K.~Yin, M.~Van De~Panne, Deeploco: Dynamic locomotion skills using hierarchical deep reinforcement learning. \emph{TOG} \textbf{36}~(4), 1--13 (2017).

\bibitem{brakel2022learning}
P.~Brakel, S.~Bohez, L.~Hasenclever, N.~Heess, K.~Bousmalis, Learning coordinated terrain-adaptive locomotion by imitating a centroidal dynamics planner, in \emph{2022 IEEE/RSJ International Conference on Intelligent Robots and Systems (IROS)} (IEEE) (2022), pp. 10335--10342.

\bibitem{mastalli2015line}
C.~Mastalli, I.~Havoutis, A.~W. Winkler, D.~G. Caldwell, C.~Semini, On-line and on-board planning and perception for quadrupedal locomotion, in \emph{2015 IEEE International Conference on Technologies for Practical Robot Applications (TePRA)} (IEEE) (2015), pp. 1--7.

\bibitem{risbourg2022real}
F.~Risbourg, \emph{et~al.}, Real-time footstep planning and control of the solo quadruped robot in 3d environments, in \emph{2022 IEEE/RSJ International Conference on Intelligent Robots and Systems (IROS)} (IEEE) (2022), pp. 12950--12956.

\bibitem{schulman2017proximal}
J.~Schulman, F.~Wolski, P.~Dhariwal, A.~Radford, O.~Klimov, Proximal policy optimization algorithms. \emph{arXiv:1707.06347}  (2017).

\bibitem{raisim}
J.~Hwangbo, J.~Lee, M.~Hutter, Per-contact iteration method for solving contact dynamics. \emph{IEEE Robot. Autom. Lett.} \textbf{3}~(2), 895--902 (2018), \url{www.raisim.com}.

\bibitem{CVAE}
D.~P. Kingma, S.~Mohamed, D.~Jimenez~Rezende, M.~Welling, Semi-supervised learning with deep generative models. \emph{Advances in neural information processing systems} \textbf{27} (2014).

\bibitem{mlp}
F.~Rosenblatt, The perceptron: a probabilistic model for information storage and organization in the brain. \emph{Psychological review} \textbf{65}~(6), 386 (1958).

\bibitem{gru}
K.~Cho, \emph{et~al.}, Learning phrase representations using RNN encoder-decoder for statistical machine translation. \emph{arXiv:1406.1078}  (2014).

\bibitem{deits2014footstep}
R.~Deits, R.~Tedrake, Footstep planning on uneven terrain with mixed-integer convex optimization, in \emph{2014 IEEE-RAS international conference on humanoid robots} (IEEE) (2014), pp. 279--286.

\end{thebibliography}
\bibliographystyle{sciencemag}

\newpage

\renewcommand{\thefigure}{S\arabic{figure}}
\renewcommand{\thetable}{S\arabic{table}}
\setcounter{figure}{0}
\setcounter{table}{0}

\section*{Supplementary Methods}
\subsection*{Network details}
\label{sup: Network details}
\subsubsection*{Actor}
Actor is a neural network responsible for controlling each of the robot's feet to step on designated target footholds. It is trained via reinforcement learning. As described in the Method section, actor's observation consists of a 167-dimensional vector comprising $O_p$, $O_h$,$O_t$, and $unObs$. Among these, $O_h$ includes history observations from 0.01, 0.02, and 0.03 seconds prior. $O_t$ contains future foothold targets for each foot, with the value of num\_target being 2. The targets for the front and rear feet represent the displacement from the base's center of mass to the target at the moment when the front\_target\_index and back\_target\_index are updated, respectively. Thus, as long as the target index remains unchanged, the target value stays constant. The $unObs$ component is a 3-dimensional vector representing the robot's linear velocity, provided by the state estimator network.
Actor's output is a 12-dimensional vector representing the joint targets, which serve as the position targets for each joint. Actor's architecture is an MLP with hidden layers of sizes [512, 128].

\subsubsection*{Critic}
Critic's observation consists of a 167-dimensional vector comprising $O_p$, $O_h$,$O_t$, and $V_{linear}$. Unlike the actor, the critic uses the actual linear velocity from the simulation rather than the estimate provided by the state estimator network. The critic shares the same MLP architecture as the actor, with hidden layers of sizes [512, 128].

\subsubsection*{State estimator network}
State estimator network is designed to estimate the robot's linear velocity and is trained via supervised learning. As described in the Method section, input is a 42-dimensional vector consisting of $O_p$ and $prev_{action}$. The network combines a GRU and an MLP, with the GRU featuring a single hidden layer of 128 dimensions. The output of the GRU serves as the input to the MLP, which then outputs the linear velocity. The MLP consists of hidden layers with sizes [64, 16].

\subsubsection*{Contact estimator}
As mentioned in the Method section, the condition for updating the target index is based on whether sufficient contact has been made. The criterion for sufficiency is that the foot has been in contact for at least 0.06 seconds. In the real experimental process, a contact estimator is utilized to estimate the contact state. The input to the contact estimator is the same as that of the Actor, and the output is a 4-dimensional vector representing the contact state for each foot. The network structure is an MLP with hidden layers of sizes [256, 64].

\subsubsection*{Conditional Variational Auto Encoder}
The encoder and decoder parts of the CVAE module are each composed of an MLP structure with hidden layers of [512,128] and [128,512], respectively. The loss function used to train the CVAE module consists of reconstruction loss and KL-divergence loss. The reconstruction loss is the MSE loss between the input to the encoder($\psi$) and the output from the decoder ( $\psi'$ ), while the KL-divergence loss measures the distance between the distribution of latent vector $z$ and a uni-normal distribution. The total loss function is formulated as a weighted sum : Total\_loss = MSE\_loss + 0.04 * KLD\_loss .

\subsection*{Components of $\psi$}
\label{sup: Components of psi}
To explain each component of $\psi$ clearly, we define the following symbols:
The vector $d_n$ represents the vector from the center of the (n-1)th stepping stone to the center of the nth stepping stone. The vector $x_n$ is the projection of $d_n$ onto the ground. The vector $Z_g$ is the vector in the direction opposite to gravity. $x_n^s$, $y_n^s$ and $z_n^s$ represent the $x$,$y$ and $z$ axes of the nth stepping stone, respectively.

Firstly, $r_n$ represents the magnitude of $x_n$ ,$\theta_n$ represents the angle between $x_{n-1}$ and $x_n$, and $\phi_n$ represents the angle between $d_n$ and $x_n$. By determining ($r_n$, $\theta_n$, $\phi_n$), we can get the center position of the nth stepping stone.
Then, we generate the n-th stepping stone at the obtained center and set its $x_n^s$ and $z_n^s$  aligned with $x_n$ and $Z_g$, respectively. 
We can obtain the final pose of the \(n\)th stepping stone by rotating it sequentially by the angles \( \Delta \text{yaw}_n \), \( x\_\text{tilt}_n \), and \( y\_\text{tilt}_n \) around the \( Z_g \), \( x_n^s \), and \( y_n^s \) axes, respectively.

\subsection*{Details of sampling and filtering footholds}
\label{sup: Details of sampling and filtering footholds}
This section details the implementation of sampling and filtering footholds in the planner module.
In the sampling process, we sequentially select $r$, $\theta$, and $\Delta_{yaw}$. At this time, the range of each component conditioned on the previous $\psi$ can be represented by a network trained through supervised learning. However, to minimize the number of networks to train and reduce time complexity, instead of using networks, we modeled the range of $r$, $\theta$, and $\Delta_{yaw}$ as polynomials with respect to the factors that have the greatest influence on each within previous $\psi$. During sampling, random sampling was then conducted within this range.

In the performance filter, six peripheral samples are selected around each left and right foot target, and these samples are located on a circle with a radius of safe\_radius on a plane parallel to the ground, centered on the target. Each of the 6 samples is chosen by recursively rotating 60 degrees clockwise from the direction of movement on the previous target. Large safe\_radius provides greater safety, but it increases the time complexity of the planning process. Through simulations and real experiments, we determined that 6 cm is a value that balances stability and time complexity well.

For the spike filter, to reduce time complexity, we reuse height measurements from the performance filter.

\subsection*{Data collection process for boundary estimator}
\label{sup: Data collection process for boundary estimator}

\begin{figure}
\includegraphics[width=\linewidth]{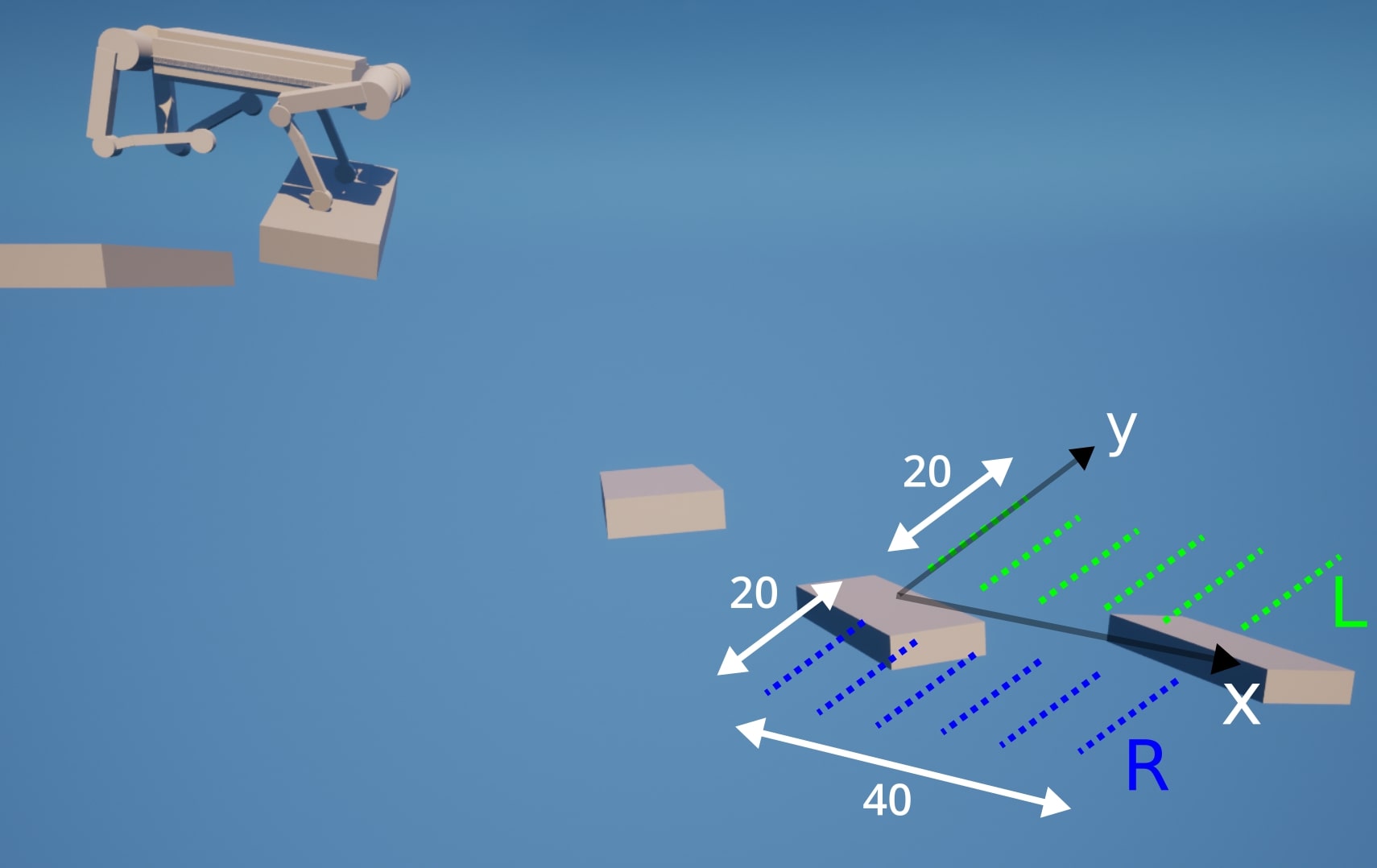}
\caption{\textbf{Grid points for collecting collision boundary data.} Showing grid points where the lowest height of the collision body was recorded.}
\label{fig:datacollection}
\end{figure}

To collect the data for training the boundary estimator, we first defined the location of grid points where the height information of the collision body would be recorded (Fig.~\ref{fig:datacollection}). 

The x-axis is defined as the ground projection component of the vector from the current target to the next target, as shown in the figure. The y-axis is a vector perpendicular to this on the ground plane. The x-coordinate of the grid points is a linear interpolation of the x-component of the current target and the x-component of the next target into 40 equal segments. The y-coordinate increases from -20 cm to 20 cm in increments of 1 cm.

As the robot moves toward the next target, each grid point records the lowest height of the collision body passing over its position. Subsequently, the lowest collision boundary for the left and right sides is determined by collecting the minimum values from the grid points along the same x-coordinate for the left and right sides relative to the y-axis, respectively. Movie S4 presents the real-time update of the lowest collision boundary as the robot moves forward.

\section*{Supplementary Results}

\subsection*{Analysis of various curricula}
\label{sup: Analysis of various curricula}
The environment in which the controller is trained consists of consecutive stepping stones, as illustrated in Fig.~\ref{fig: mapgenerator}A. For robust high-speed locomotion, it is necessary to incorporate control strategies that account for deceleration and acceleration. To achieve this, the controller must be informed about future targets. Therefore, we provide the controller with information about the next two targets for both the front and rear feet as input.
As a result, the robot's movement while traversing the (n-2)th stepping stone is influenced by both the (n-1)th target and the nth target. Therefore, to create feasible environments, it is necessary to recursively determine the next target in a way that makes the previous two targets feasible. In other words, to construct a feasible environment, when generating $\psi_n$, one must consider $\psi_{n-2}$, $\psi_{n-1}$, and $T_{last}$. 
To represent such an environment, a 19-dimensional high-dimensional space of \{$\psi_{n-2}$,$\psi_{n-1}$,$\psi_n$,$T_{last}$\} is required.

\begin{figure}
\includegraphics[width=\linewidth]{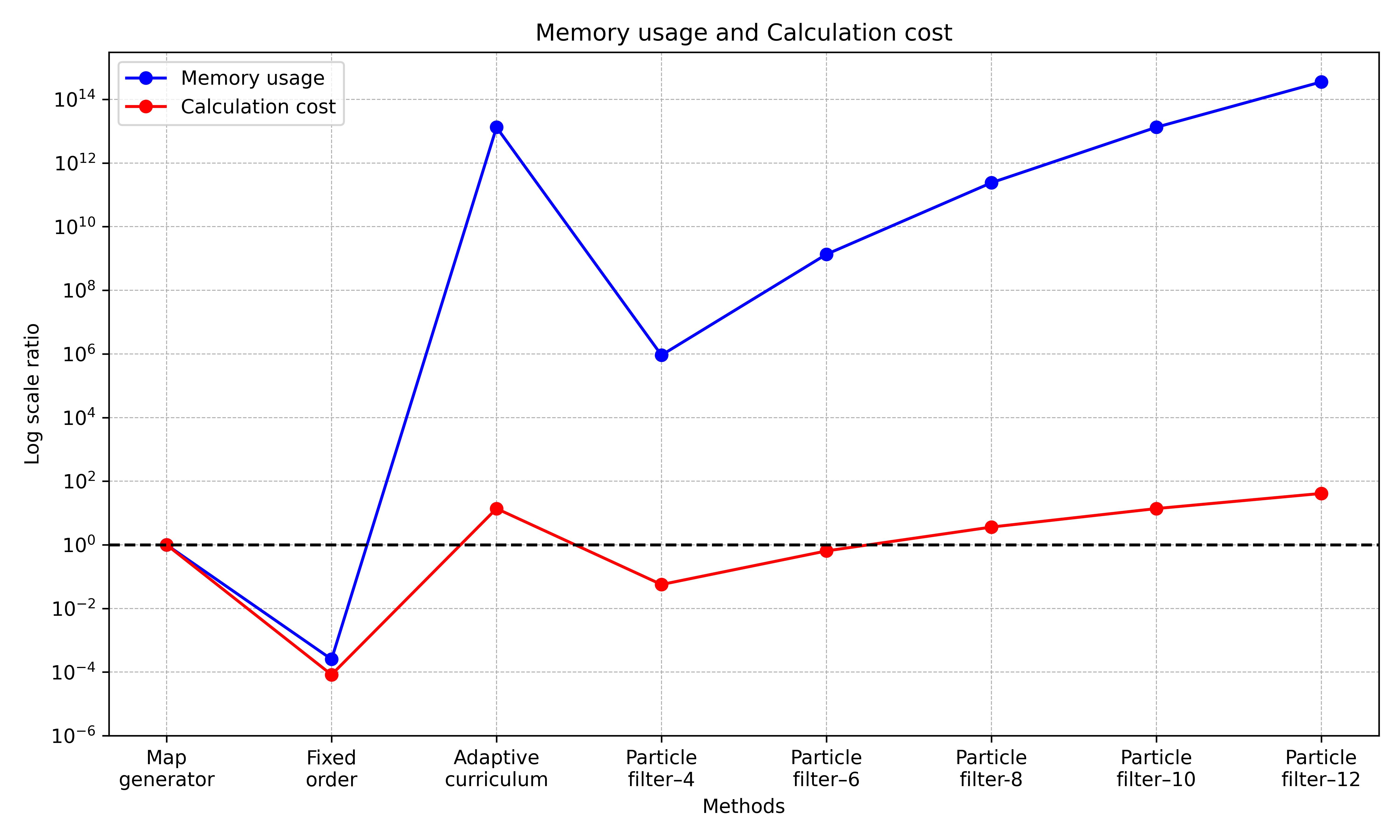}
\caption{\textbf{Memory usage and Calculation cost.} Memory usage and calculation cost compared to the map generator were measured from various curriculum methods and settings. The method labeled as "Particle filter-n" indicates the particle filtering-based method with the number of particles for one dimension set to n.}
\label{fig:curricula}
\end{figure}

We adopted 3 curriculum methods as a baseline to validate the proposed map generator method.
Baseline 1 is the commonly used fixed-order curriculum method \cite{xie2020allsteps} that advances evenly through parameter space. 
Baseline 2 is an adaptive curriculum method proposed in \cite{xie2020allsteps}, which explores evenly divided parameter space at different paces depending on the map's difficulty. 
Baseline 3 is an automatic terrain curriculum proposed in \cite{lee2020}. It adapts particle filtering to maintain an appropriate level of map difficulty. 
We measured the memory usage and the number of computations required on both baselines and the proposed method, as shown in Fig.~\ref{fig:curricula}.

\begin{figure}
\includegraphics[width=\linewidth]{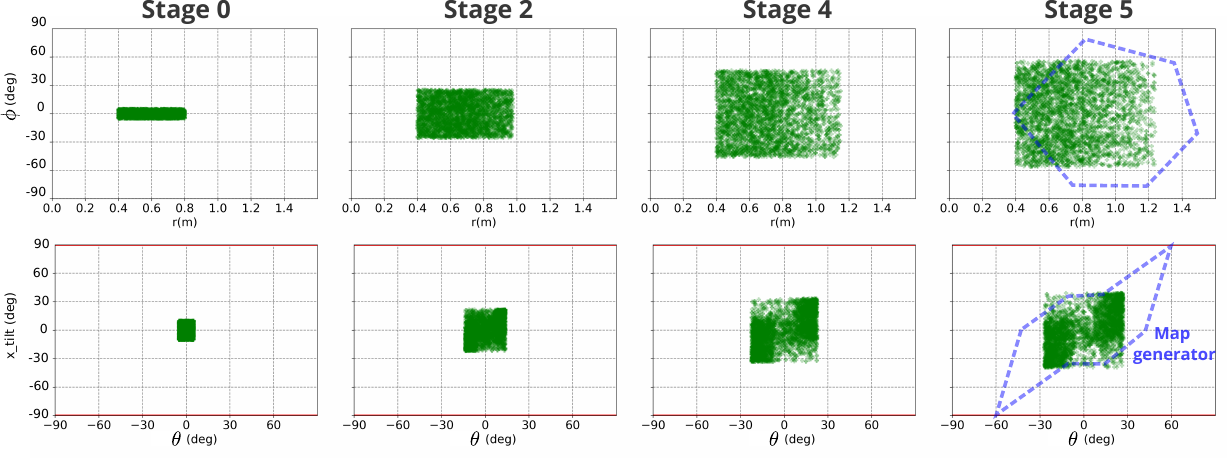}
\caption{\textbf{Evolution of $\psi$ distribution with a fixed-order curriculum.} The distribution of the $\psi$ parameters across different stages when using the fixed-order curriculum is shown. In stage 5, the blue dashed line represents the boundary of the $\psi$ distribution at the final stage of training with the map generator.}
\label{fig:fixed evolve}
\end{figure}

Baseline 1 has the advantage of being simple to implement, requiring minimal memory, and needing very few computations to generate parameters, as evidenced in Fig.~\ref{fig:curricula}. However, since this method merely expands the range of each parameter in a fixed order, it struggles to represent the complex distributions of feasible parameters in high-difficulty environments.
When training the controller using baseline 1, we observed that it evolved well until moderate-difficulty environments. Fig.~\ref{fig:fixed evolve} shows the evolution of parameter distribution as the stage is updated. However, from stage 5 onward, when the distribution of infeasible parameters became too prevalent, the controller failed to evolve further. In the map generated using this parameter distribution, the controller trained with the baseline 1 method can only overcome an average of 4.3 out of 10 stepping stones.

Baseline 2 evenly divides the parameter space into multiple cells, and maps are constructed with a probability depending on the weight stored in each cell. It can better represent complex parameter distribution than Baseline1. However, if the cells are not divided into a sufficient number, this method may fail to capture the intricate distribution of feasible parameters. As the dimension increases, the number of cells required to represent the distribution grows exponentially, leading to a significant increase in memory usage. It required significant memory usage (approximately $10^{13}$ times more memory than our method), which made it challenging to implement and apply it directly.

Similarly, baseline 3 also has the advantage of effectively representing complex distributions, but as the dimension of the space increases, the need for a large number of particles results in substantial memory usage. fig.~\ref{fig:curricula} illustrates the computational load and memory requirements of the Particle filter-n method, depending on the number of particles (n) comprising one dimension of the parameter space. It also required significant memory usage, making it challenging to implement and apply directly.

In contrast, the proposed map generator method uses a generative model capable of considering conditions. Instead of simply storing weight values for each (condition, next target) pair like in baselines 2 and 3, it approximates the relationship between condition ($\psi_{n-2}$,$\psi_{n-1}$,$T_{last}$) and feasible next target parameter($\psi_n$), allowing it to represent high-dimensional spaces while using significantly less memory.


\clearpage

\clearpage
\begin{table}[ht]
    \centering
    \caption{PPO algorithm parameters}
    \vspace{4pt}
    \label{tab:ppo_parameters}
    \begin{tabular}{|l|c|}
        \noalign{\hrule height 0.4pt} 
        \textbf{Parameter} & \textbf{Value} \\
        \noalign{\hrule height 3pt} 
        num\_learning\_epoch & 16 \\
        \noalign{\hrule height 0.4pt}
        num\_learning\_epoch (map generator) & 12 \\
        \noalign{\hrule height 0.4pt}
        $\gamma$ & 0.9950 \\
        \noalign{\hrule height 0.4pt}
        $\lambda$ & 0.9500 \\
        \noalign{\hrule height 0.4pt}
        learning rate & 0.0002 \\
        \noalign{\hrule height 0.4pt}
        learning rate (map generator) & 0.0001 \\
        \noalign{\hrule height 0.4pt}
        max grad norm & 0.5000 \\
        \noalign{\hrule height 0.4pt}
    \end{tabular}
\end{table}

\begin{table}[ht]
\centering
\caption{Domain randomization}
\vspace{4pt} 
\label{tab:domain_randomization}
\renewcommand{\arraystretch}{0.7}
\begin{tabular}{|c|c|}
\hline
\textbf{\begin{tabular}[c]{@{}c@{}}Randomization\\ Type\end{tabular}} & \textbf{Range} \\
\noalign{\hrule height 3pt}

\rule{0pt}{2.5ex}\textbf{PD gain} & \rule{0pt}{2.5ex}-10$\sim$10 \% \\
\hline
\rule{0pt}{2.5ex}\textbf{\begin{tabular}[c]{@{}c@{}}Control time\\ delay\end{tabular}} & \rule{0pt}{2.5ex}0$\sim$30 \% \\
\hline
\rule{0pt}{2.5ex}\textbf{Observation} & \rule{0pt}{2.5ex}-10$\sim$10 \% \\
\hline
\rule{0pt}{2.5ex}\textbf{Initial state} & \rule{0pt}{2.5ex}-10$\sim$10 \% \\
\hline
\rule{0pt}{2.5ex}\textbf{\begin{tabular}[c]{@{}c@{}}History\\ observation\end{tabular}} & \rule{0pt}{2.5ex}-10$\sim$10 \% \\
\hline
\rule{0pt}{2.5ex}\textbf{Foot mass} & \rule{0pt}{2.5ex}-7$\sim$7 \% \\
\hline
\rule{0pt}{2.5ex}\textbf{Base property} & \rule{0pt}{2.5ex}\begin{tabular}[c]{@{}c@{}}mass: 0$\sim$40 \%\\ COM\_position: (0$\sim$15 cm, 0$\sim$2 cm, 0$\sim$2 cm)\\ inertia: 0$\sim$50 \%\end{tabular} \\
\hline
\rule{0pt}{2.5ex}\textbf{\begin{tabular}[c]{@{}c@{}}Friction\\ coefficient\end{tabular}} & \rule{0pt}{2.5ex}0.4-1.0 \\
\hline
\end{tabular}
\end{table}

\begin{table}[ht]
\centering
\caption{Initial curriculum parameter}
\vspace{4pt}
\label{tab:linear_curricula}
\renewcommand{\arraystretch}{0.7}
\begin{tabular}{|l|l|l|l|l|l|}
\hline
\textbf{Parameter} & \textbf{Stage 0} & \textbf{Stage 1} & \textbf{Stage 2} & \textbf{Stage 3} & \textbf{Stage 4} \\
\noalign{\hrule height 3pt}

\rule{0pt}{2.5ex}$r_{low}$ (m) & 0.4 & 0.4 & 0.4 & 0.4 & 0.4 \\
\hline
\rule{0pt}{2.5ex}$r_{high}$ (m) & 0.8 & 0.875 & 0.95 & 1.025 & 1.1 \\
\hline
\rule{0pt}{2.5ex}$\theta$ (deg) & 5 & 8.75 & 12.5 & 16.25 & 20 \\
\hline
\rule{0pt}{2.5ex}$\phi$ (deg) & 5 & 13.75 & 22.5 & 31.25 & 40 \\
\hline
\rule{0pt}{2.5ex}$\Delta_{yaw}$ (deg) & 0 & 2.5 & 5 & 7.5 & 10 \\
\hline
\rule{0pt}{2.5ex}$x_{tilt}$ (deg) & 10 & 15 & 20 & 25 & 30 \\
\hline
\rule{0pt}{2.5ex}$y_{tilt}$ (deg) & 5 & 6.25 & 7.5 & 8.75 & 10 \\
\hline
\end{tabular}
\end{table}

\begin{table}[t]
    \centering
    \caption{Reward functions and coefficients}
    \vspace{4pt}
    \label{tab:reward_function}
    \renewcommand{\arraystretch}{1.2}
    \small
    \begin{tabular}{|c|l|l|c|}
        \hline
        \textbf{Category} & \textbf{Reward} & \textbf{Equation} & \textbf{Coefficient} \\
        \hline

        \textbf{Target} & Target\_sparse &
        \begin{tabular}[l]{@{}l@{}}
        $R_{ts} = \begin{cases}
            k_{ts1}*{k_{ts2}}^{count_{front}} e^{-k_{ts3}(d_{fl}+d_{fr})}, & \text{if } C_0 \text{ or } C_1 \\
            k_{ts1}*{k_{ts2}}^{count_{back}} e^{-k_{ts3}(d_{rl}+d_{rr})}, & \text{if } C_2 \text{ or } C_3 \\
            0, & \text{else}
        \end{cases}$
        \end{tabular}
        & \begin{tabular}[c]{@{}c@{}}$k_{ts1}:$ 9.4\\ $k_{ts2}:$ 0.97\\ $k_{ts3}:$ 6\end{tabular} \\
        
        & Target\_dense & $R_{td} = k_{td}(0.8-(d_f+d_r))$ & $k_{td}:$ 0.30 \\

        & Target\_last &
        $
        R_{tl} = \begin{cases}
            k_{tl}, & \text{if } T_{last} \land C_0 \land C_1 \land C_2 \land C_3 \\
            0, & \text{else}
        \end{cases}
        $
        & $k_{tl}:$ 5 \\
        \hline

        \textbf{Style} & Torque & $R_{\tau} = k_{\tau} cf (|\tau_{max}|^3 - 35^3)$ & $k_{\tau}:$ -2.5e-05 \\

        & Slip &
        $R_{slip} = \sum_{i=0}^{3} \begin{cases}
            k_{s} cf |V_{xy_i}|, & \text{if } C_i \\
            0, & \text{else}
        \end{cases}$
        & $k_{s}:$ -0.03 \\

        & \begin{tabular}[l]{@{}l@{}}Foot gather\\longitudinal\end{tabular} &
        $R_{longi} = \begin{cases}
            k_{longi} \max(d_{fb}, 0.3), & \text{if front\_target\_updated} \\
            0, & \text{else}
        \end{cases}$
        & $k_{longi}:$ -1 \\

        & \begin{tabular}[l]{@{}l@{}}Foot gather\\lateral\end{tabular} &
        $R_{lat} = \sum_{i=0}^{1} \begin{cases}
            k_{lat} \min(d_{lr_i}, 0.25), & \text{if } C_{2i} \land C_{2i+1} \\
            0, & \text{else}
        \end{cases}$
        & $k_{lat}:$ -1 \\

        & Bound &
        $R_{bound} = \begin{cases}
            k_{bound}, & \text{if } (C_0=C_1) \land (C_2=C_3) \land \lnot(C_0=1 \land C_2=1) \\
            0, & \text{else}
        \end{cases}$
        & $k_{bound}:$ 1 \\

        & Joint velocity & $R_{jv} = k_{jv} cf |w|$ & $k_{jv}:$ -5e-05 \\

        & Stop &
        $R_{stop} = \begin{cases}
            k_{stop}(|V_x| + |\omega_{pitch}|/3 - 2), & \text{if } T_{last} \\
            0, & \text{else}
        \end{cases}$
        & $k_{stop}:$ -0.3 \\

        & Impact &
        $R_{impact} = \sum_{i=0}^{3} \begin{cases}
            k_{impact} cf v_{\perp}^2, & \text{if i-th foot touchdown} \\
            0, & \text{else}
        \end{cases}$
        & $k_{impact}:$ -2 \\

        & Smooth &
        $R_{smooth} = k_{smooth} \max(0, cf-0.8) |(T_k-T_{k-1})-(T_{k-1}-T_{k-2})|$
        & $k_{smooth}:$ -2 \\
        \hline

        \textbf{Constraint} & Joint limit &
        $R_{limit} = \sum_{i=0}^{3} k_{limit} cf \min(e^{(\theta_{ki}+1.24)^4}-1, 20)$
        & $k_{limit}:$ -0.20 \\
        \hline
    \end{tabular}
\end{table}

\begin{table}[ht]
    \centering
    \renewcommand{\arraystretch}{1.0}
    \begin{tabular}{l|l}
        \hline
        \textbf{Symbol} & \textbf{Description} \\
        \noalign{\hrule height 1.2pt}

        $count_{front}, count_{back}$ & The number of simulation steps since front/rear foot touchdown \\
        $d_{fl}, d_{fr}, d_{rl}, d_{rr}$ & Distance between each target and the foot \\
        $d_f, d_r$ & Distance from center of front/rear feet to center of recent target \\
        $T_{last}$ & A boolean indicating whether target\_index is the last value \\
        $C_i$ & Contact state of the i-th foot (0 for swing, 1 for stance) \\
        $cf$ & Curriculum factor \\
        $\tau_{max}$ & Maximum value of size of joint torque \\
        $V_{xy_i}$ & Horizontal velocity of i-th foot \\
        $d_{fb}$ & Distance between center of front feet and center of rear feet \\
        $d_{lr_i}$ & Distance between left and right foot (front (i=0), rear (i=1)) \\
        $w$ & 12-dimensional joint velocity vector \\
        $V_{\perp}$ & Foot velocity perpendicular to the target plane \\
        $\theta_{ki}$ & Angle of i-th knee joint \\
        $T_k$ & Joint target at control time step k \\
    \end{tabular}
\end{table}

\begin{table}[thpb]
    \centering
    \caption{Planning costs}
    \label{tab:planning_cost}
    \renewcommand{\arraystretch}{1.0}
    \small
    \begin{tabular}{l|l|l}
        \hline
        \textbf{Cost} & \textbf{Equation} & \textbf{Coefficient} \\
        \noalign{\hrule height 1.2pt}

        Survive\_cost &
        $
        C_{survive} = 
        \begin{cases}
            0 & \text{if termination} \\
            k_{survive} & \text{else}
        \end{cases}
        $
        & $k_{survive}:$ -10000 \\

        Distance\_cost &
        $C_{distance} = k_{distance} \times \min(10, d_g)$
        & $k_{distance}:$ 1 \\

        Direction\_cost &
        $C_{direction} = k_{direction} \times \theta_g$
        & $k_{direction}:$ 0.0333 \\

        Elevation\_cost &
        $
        C_{elevation} = k_{elevation} \sum_{i=k}^{k+\text{predict\_horizon}-1} |T_{i+1} - T_i|
        $
        & $k_{elevation}:$ 1 \\
    \end{tabular}
\end{table}

\begin{table}[ht]
    \centering
    \renewcommand{\arraystretch}{1.0}
    \small
    \begin{tabular}{l|l}
        \hline
        \textbf{Symbol} & \textbf{Description} \\
        \noalign{\hrule height 1.2pt}

        $d_g$ & Distance from goal to the final position of the robot at roll out \\
        $\theta_g$ & Angle between the displacement vector and the vector from initial position to the goal \\
        $T_i$ & Height of foothold target at target index $i$ \\
    \end{tabular}
\end{table}
\clearpage
\newpage

\begin{algorithm}
\caption{Adversarial training process}
\label{alg:adversarial training}
\begin{algorithmic}[0]
\State $\alpha \gets 0.7$ \Comment{Parameter for adjusting map difficulty}

\For{\(0 \leq \text{update} \leq \text{max\_update}\)}
    \State map\_generator.generate($\alpha$) \Comment{Generate $\psi$s and provide it as environment}
    \State env.roll\_out() \Comment{Roll out for max\_time (4.2s)}
    \State avg\_performance $\gets$ env.check\_performance()
    \State actor.update() \Comment{Train actor}
    \If{update \% update\_period == 0 \textbf{and} avg\_performance $>$ 9.3}
        \State feasible\_param $\gets$ env.get\_feasible\_param() 
        \State map\_generator.retrain(feasible\_param)
        \State $\alpha \gets 0.7$
        \State map\_generator.generate($\alpha$) \Comment{Generate map parameters from retrained one}
        \State env.roll\_out()
        \State avg\_performance $\gets$ env.check\_performance()
        \While{avg\_performance $<$ 9.15} \Comment{Reduce map difficulty until desired difficulty}
            \State $\alpha \gets \alpha - 0.02$ 
            \State map\_generator.generate($\alpha$)
            \State env.roll\_out()
            \State avg\_performance $\gets$ env.check\_performance()
        \EndWhile
    \EndIf
\EndFor
\end{algorithmic}
\end{algorithm}

\begin{algorithm}
\caption{Deployment process}
\label{alg:Deployment}
\begin{algorithmic}[0]
\State goal\_reached $\gets$ \textbf{False}
\For{\(0 \leq \text{step} \leq \text{total\_steps} - 1\)}
    \State tracker.control() \Comment{Runs at 100Hz}
    \State front\_index\_updated $\gets$ tracker.check\_update()
    \If{front\_index\_updated \textbf{and not} goal\_reached} \Comment{Run in detached thread}
        \State planner.synchronize() \Comment{Synchronize with tracker's situation}
        \State planner.sample\_foothold\_plans()
        \State planner.roll\_out()
        \State planner.evaluate\_cost()
        \State best\_plan $\gets$ planner.get\_best\_plan()
        \State goal\_reached $\gets$ planner.check\_goal\_reached()
        \State tracker.update\_foothold\_target(best\_plan)
    \EndIf
\EndFor
\end{algorithmic}
\end{algorithm}

\end{document}